\pdfoutput=1
\documentclass[10pt,twocolumn,letterpaper]{article}

\usepackage[pagenumbers]{cvpr} 

\definecolor{cvprblue}{rgb}{0.21,0.49,0.74}
\usepackage[pagebackref,breaklinks,colorlinks,allcolors=cvprblue]{hyperref}
\usepackage{graphicx}
\usepackage{wrapfig}
\usepackage{multirow}
\usepackage{booktabs} 
\usepackage{caption}
\usepackage{pifont}
\usepackage{bbding}
\usepackage{algorithm}
\usepackage{booktabs}
\usepackage{algpseudocode}
\usepackage[accsupp]{axessibility}  
\def\paperID{12959} 
\def\confName{CVPR}
\def\confYear{2025}

\title{PCM : Picard Consistency Model for Fast Parallel Sampling of Diffusion Models}

\author{
Junhyuk So$^{1}$ \quad
Jiwoong Shin$^{1}$ \quad
Chaeyeon Jang$^{1}$ \quad
Eunhyeok Park$^{1,2}$ \\
$^{1}$Department of Computer Science and Engineering, POSTECH\\
$^{2}$Graduate School of Artificial Intelligence, POSTECH\\
{\tt\small \{junhyukso,jwshin0610, jcy2749, eh.park\}@postech.ac.kr}
}

\begin{document}
\begingroup

\maketitle

\begin{abstract}
Recently, diffusion models have achieved significant advances in vision, text, and robotics. However, they still face slow generation speeds due to sequential denoising processes. To address this, a parallel sampling method based on Picard iteration was introduced, effectively reducing sequential steps while ensuring exact convergence to the original output. Nonetheless, Picard iteration does not guarantee faster convergence, which can still result in slow generation in practice. In this work, we propose a new parallelization scheme, the Picard Consistency Model (PCM), which significantly reduces the number of generation steps in Picard iteration. Inspired by the consistency model, PCM is directly trained to predict the fixed-point solution, or the final output, at any stage of the convergence trajectory. Additionally, we introduce a new concept called model switching, which addresses PCM’s limitations and ensures exact convergence. Extensive experiments demonstrate that PCM achieves up to a 2.71x speedup over sequential sampling and a 1.77x speedup over Picard iteration across various tasks, including image generation and robotic control.
\end{abstract}

\renewcommand\thefootnote{}\footnote{This paper was accepted to the IEEE/CVF Conference on Computer Vision and Pattern Recognition (CVPR) 2025.}
\addtocounter{footnote}{-1}
\endgroup    
\section{Introduction}
\label{sec:introduction}

Recently, diffusion probabilistic models \cite{ddpm} have achieved outstanding performance across diverse generative tasks in vision \cite{ldm,sdxl,guided}, audio \cite{audioldm,diffwave,makeaudio}, text \cite{textdiffusionicml,tiffusion,austin2021structured}, robotics \cite{diffusionpolicy, dallebot}, and bioinformatics \cite{dnadiffusion,RFDiffusion}. Their iterative denoising process enables high-quality outputs by capturing fine details and diversity, making them versatile across applications. However, this process has a notable drawback: slow generation speed. For example, early diffusion models like DDPM \cite{ddpm} require 1,000 denoising steps, each involving intensive computation, resulting in high latency.

To enable faster inference, various approaches have been proposed; however, most achieve speedups at the expense of quality degradation or inconsistent convergence. For instance, ODE-based solvers like DDIM \cite{ddim} often trade off image quality for fewer sampling steps. Studies like the Consistency Model \cite{song2023consistency} and Progressive Distillation \cite{salimans2022progressive} use distillation to reduce sampling steps, but the final output may differ from the original model due to modified weights.

In this work, we aim to explore a different question: can increased parallelism lead to faster convergence? While adding parallelism can nearly linearly enhance throughput, achieving low latency in sequential generative models demands innovation at the mathematical foundation level. Recently, ParaDiGMS \cite{shih2024parallel} tackled this challenge with a parallel sampling method based on Picard iteration, reducing the sequential steps in diffusion sampling by performing more computations in parallel. Picard iteration ensures convergence to the exact output, guaranteeing the same output quality as original model; however, it does not ensure fast convergence, which may result in slower generation speeds in practice.

In this study, we introduce the Picard Consistency Model (PCM) to accelerate the convergence speed of Picard iteration. Inspired by the Consistency Model \cite{song2023consistency}, PCM is trained to directly predict the fixed-point solution at any point along the Picard iteration trajectory. In addition, our novel idea, model mixing, ensures exact convergence to the original output even with PCM's acceleration. Our extensive analysis reveals that PCM with model mixing gives a 2.71x speedup over sequential denoising and 1.77x speedup over Picard iteration, with guaranteeing exact convergence in diverse image generation and robotic applications. 

\section{Preliminary}
\label{sec:preliminary}
\begin{figure*}[t!]
    \centering
        \includegraphics[width=0.98\linewidth]{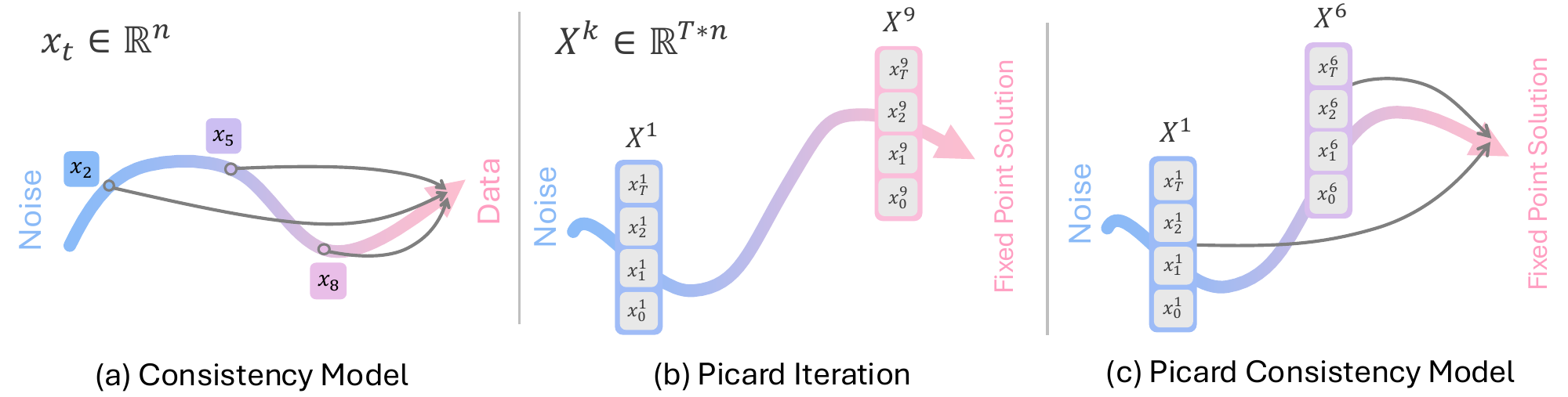}
    \caption{\textbf{Visual comparison} of (a) Consistency Model \cite{song2023consistency}, (b) Picard iteration, and (c) our Picard Consistency Model (PCM). (a) \cite{song2023consistency} is trained to directly predict final output $x_T$ from any point $x_t\in\mathbb{R}^n$ in denoising trajectory. (b) Picard iteration forms a denoising trajectory along tensors $X\in\mathbb{R}^{T*n}$. (c) Inspired by \cite{song2023consistency}, we train PCM to predict final point $X^*$ from any intermediate step in the Picard trajectory.}
    \label{fig:viz_comp}
\end{figure*}

\subsection{Diffusion Models}

The Diffusion Probabilistic Model (DPM) was first introduced in the Denoising Diffusion Probabilistic Models \cite{ddpm} (DDPM) . The latent variable of diffusion model, ${x_t (t\le T)}$ is defined through following diffusion forward process $q(\cdot)$.

\begin{equation}
    q(x_t|x_T) = N(x_t; \sqrt{(\alpha(t))}x_T, (1-\alpha(t))I)
\end{equation}

Here, $\alpha(t)$ is scalar function that setted to $\alpha(0)=0$, transforming $q(x_t)$ to standard gaussian $\mathcal{N}(0,I)$ with $t \rightarrow 0$. Please note that we denote $p(x_T)$ as a data distribution. The reverse process, $p_\theta(x_{t+1}|x_{t})$, is also parameterized by gaussian process with predicted mean and time-dependant predefined variance $\sigma^2_t$.

\begin{equation}
    p_\theta(x_{t+1}|x_{t}) = N(x_{t+1}; \mu_\theta(x_{t}), \sigma_{t}^2I)
\end{equation}

As shown in \cite{song2020score}, the notable property of the diffusion model is that the forward process can also be expressed in the form of a Stochastic Differential Equation (SDE). The SDE version of diffusion forward process is as follows :

\begin{equation}
    dx_t = f(t)x_tdt+g(t)d\mathcal{W},    x_T\sim q(x_T).
\end{equation}

Here, $\mathcal{W}$ is the standard Wiener process and $f(t),g(t)$ is the drift and diffusion coefficient function that chosen to match $q(x_t|x_T)$. Similarly, the reverse process can also be represented in the form of SDE. 

\begin{equation}
    dx_t = [f(t)x_t  - g^2(t)\nabla_{x_t}\log{q_t(x)}]dt + g(t)d\mathcal{W},
\end{equation}
where now the term $\nabla_{x_t}\log{q_t(x)}$ is approximated by our neural network $S_\theta(\cdot)$, with learnable weights of $\theta$.
As known in \cite{song2020score}, these SDE has their corresponding \textit{probability flow ODE}, which is given as follows:

\begin{equation}
    dx_t = [f(t)x_t  - g^2(t)\nabla_{x_t}\log{q_t(x)}]dt \label{eq:5}.
\end{equation}

This ODE, with the randomness term ($g(t)d\mathcal{W}$) removed, generally has slightly lower generation performance than SDE but enables inference with very few discretization steps. Now we will focus on how to effectively solve this oridinary diffrental equation with parallel computing.

\begin{figure*}[t!]
    \centering
        \includegraphics[width=0.9\linewidth]{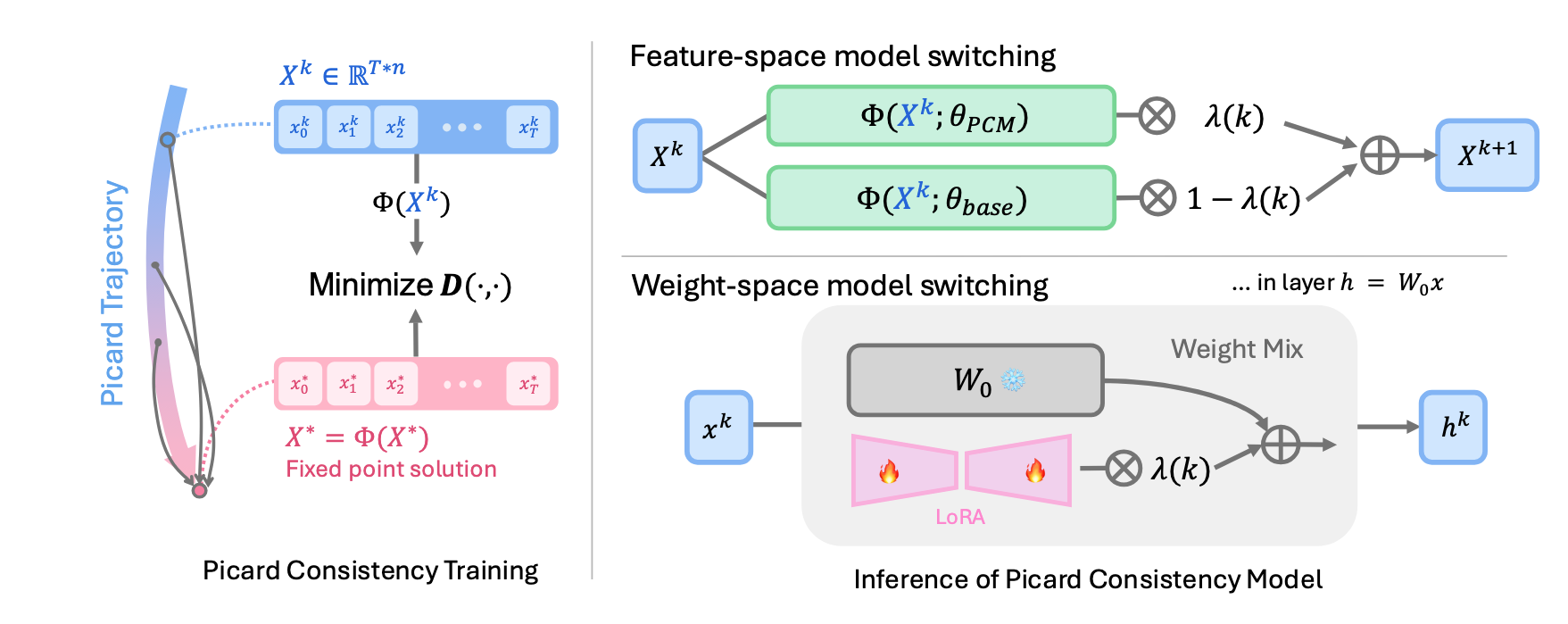}
    \caption{\textbf{Overview of our method.} Our goal is to accelerate the Picard iteration process, \( X^{k+1} \leftarrow \Phi(X^k) \), by training with a loss function that minimizes the distance between a random point on the Picard trajectory and the fixed-point solution \( X^* \). During inference, to preserve the convergence properties of the Picard iteration, we smoothly transition from our trained model \( \theta_{PCM} \) to the original model \( \theta_{base} \) using a scheduling function \( \lambda(k) \) in feature space or weight space through LoRA.}
    \label{fig:main_fig}
\end{figure*}

\subsection{Picard Iteration}
Another well-explored way to solve differential equation is transforming it as a Initial Value Problem (IVP). 
The initial value problem of differential equation, such as Eq. \ref{eq:5}, can be generally defined as 
\begin{equation}
    x'(t) = F(t, x(t)), x(0) = x_0,
\end{equation}
where  $x'$ denotes the differential($dx/dt$) and $x$ and $F$ are function that $x : \mathbb{R} \rightarrow \mathbb{R}^n$ , $F : \mathbb{R}\times\mathbb{R}^n \rightarrow \mathbb{R}^n$ and $x_0$ is the initial value. The equivalent integral form of this IVP is

\begin{equation}
    x(t) = x(0) + \int_0^tF(x(s),s)ds.
\end{equation}

Intuitively, evaluating this integral for x(t) requires the values of $x(t)$ over the interval $[0,t-\delta t)$, necessitating sequential evaluation. To introduce parallelism for solving this equation, we define the integral operator $\Phi$ on $x$. 
\newtheorem{definition}{Definition}
\newtheorem{theorem}{Theorem}

\begin{definition}[Integral Operator]\label{def:io}
    For any integral form of IVP, we can define the integral operator,$\Phi$ as
    \[
    \Phi(x;F) = x(0) + \int_0^t F(x(s), s) \, ds ,
    \]
    where $\Phi : \mathcal{X} \rightarrow \mathcal{X}$ and $\mathcal{X}$ is the function space of $x$.
\end{definition}

In the context of diffusion models, we can interpret \( F(\cdot) \) as a single evaluation of time-conditioned diffusion neural network, \( x(t) \) as the denoising trajectory at time \( t \), and \( \Phi(\cdot) \) as the inference process of a differential equation solver (e.g., DDIM \cite{ddim}). As shown in Def. \ref{def:io}, the \textit{inference} of \( \Phi \) does not necessarily need to be sequential in \( t \), because now we assume access to the entire trajectory of \( x(t) \), even if it is not the exact solution of the differential equation. Based on this definition, the Picard iteration algorithm provides a new way to solve this IVP in parallel, leveraging the well-known Picard-Lindelöf theorem 
\cite{coddington1956theory}.

\begin{theorem}[Picard–Lindelöf Theorem \cite{coddington1956theory}]
    Let $F$ be continuous in $t$ and Lipschitz continuous in $x$. Then $\Phi$ has a unique fixed-point solution and is a \textbf{contraction mapping}, satisfying $d(\Phi(X),\Phi(Y)) \le q \cdot\, d(X,Y)$, where $(X, d)$ is a complete metric space and $q \in [0,1)$.
\label{thm:picard}
\end{theorem}

Based on these theorems , the  picard iteration algorithm is derived to find the fixed-point solution of IVP through the recursive inference on $\Phi$.
\begin{equation}
    \textbf{picard iteration : } x^{k+1} \leftarrow  \Phi(x^k) , x(0)=x_0
\end{equation}

Starting with any initial \( x_0 \), this iteration converges to a unique fixed-point solution \( x^* = \Phi(x^*) \) as \( k \to \infty \), as long as that \( F \) satisfies the mild assumptions in Theorem \ref{thm:picard}.

In practice, since the actual inference process of diffusion models is discretized in time, we can reinterpret this algorithm where $\Phi$ is now defined on tensors $X \in \mathbb{R}^{T \times n}$, with $T$ denoting the total number of discretization time steps. ParaDiGMS\cite{shih2024parallel} proposed the following discretized Picard iteration for discrete-time diffusion models:

\begin{align}
    X^{k+1} &\leftarrow \Phi(X^k) \\
    x^{k+1}_t &\leftarrow x_0^k + \frac{1}{T} \sum_{i=0}^{t-1} s_{\theta}(x_i^k, i/T)
\end{align}

Here, $x^k_i \in \mathbb{R}^n$ represents the denoising result at the $i$-th time step during the $k$-th picard iteration. They observed that $k$ is typically smaller than \( T \), enabling practical and efficient parallel sampling of diffusion models.

\section{Method}
\label{sec:method}

The mathematical foundation of Picard iteration supports convergence, but as shown in studies on numerical analysis \cite{nevanlinna1990linear, pollock2019anderson, hutzenthaler2021speed}, its convergence rate is often undefined and tends to be slow for large-scale problems, limiting practical speedup. To address this, we propose a novel method, Picard Consistency Training (PCT), to accelerate Picard iterations in diffusion models while preserving exact convergence properties. 

\subsection{Picard Consistency Training}

Our approach is inspired by the intuitive observation that even minimal tuning can significantly enhance convergence. As shown in several diffusion distillation studies \cite{song2023consistency, salimans2022progressive, LCM, guided}, diffusion models have the capability to predict multiple steps ahead in their denoising trajectory with proper guidance from a pretrained teacher model. For instance, Consistency Models \cite{song2023consistency} are trained to reach their final output $x_T$ directly from any point along the denoising trajectory. Similarly, while the denoising process in diffusion models forms a trajectory along a single tensor $x \in \mathbb{R}^n$, the Picard iteration can be viewed as forming a trajectory along tensors $X \in \mathbb{R}^{T*n}$. Building on this insight, we propose our Picard Consistency Model(PCM) in the style of Consistency Models: training the model to predict its final fixed-point solution from any iteration along the Picard trajectory. A visual comparison is presented in Fig. \ref{fig:viz_comp}.

\textbf{Picard Trajectory Dataset Generation.} First, we collect the trajectory of the Picard iterations from the original model to make our \textit{Picard Trajectory Dataset}. Specifically, to generate \( N \) samples for our dataset \( \mathcal{D} \), we prepare our pretrained diffusion model \( \theta_{\text{base}} \) and differential equation solver \( \Phi(\cdot) \). Then, we sample an initial noise \( x_0 \in \mathbb{R}^n \) from a standard Gaussian distribution \( \mathcal{N}(0, I) \) and run the Picard iteration with \( \theta_{\text{base}} \) for a total of \( K \) iterations. During each iteration of the Picard process, we store the initial noise \( x_0 \) and the current result of each Picard iteration \( X^k \in \mathbb{R}^{T \times n} \). We provide the detailed algorithm for generating the trajectory dataset in Algorithm \ref{algo:1}.

\textbf{Loss function.} For training, we randomly sample data \( X \) from our trajectory dataset \( \mathcal{D} \) and choose a random index \( k \) from a uniform distribution \( \mathcal{U}[0, K-1] \). We then select two tensors: \( X^k \), a random point on the trajectory, and \( X^K = X^* \), the solution fixed point. Initialized with the pretrained diffusion model \( \theta_{\text{base}} \) used to generate the trajectory dataset, we train our PCM \( \theta_{\text{PCM}} \) to directly predict the Picard solution \( X^* \) by using \( X^k \) as the input of a single Picard iteration. The model is trained by minimizing the following loss function:

\begin{equation}
    \mathcal{L} = \mathbb{E}_{X \sim \mathcal{D}, \, k \sim \mathcal{U}[0, K-1]} \, \alpha(k) \, D(X^{*}, \Phi(X^k; \theta_{\text{PCM}})), \label{eq:11}
\end{equation}
where \( D(\cdot, \cdot) \) is the metric function used to measure the distance between two data distributions (e.g., L2). One important aspect to note is the weighting function \( \alpha(k) \), which is used to mitigate the noise variance gap between \( X^k \) and \( X^* \). Intuitively, since the early stages (smaller \( k \)) of the Picard iteration are farther from the solution point, the images tend to be noisier compared to later stages. We use \( \alpha(k) = \frac{1}{\sqrt{\text{Var}(k)}} \), where \( \text{Var}(k) \) is the predefined variance schedule of the original diffusion model.

\subsection{Feature-space Model Switching}

\begin{figure}[t!]
    \centering
        \includegraphics[width=0.97\linewidth]{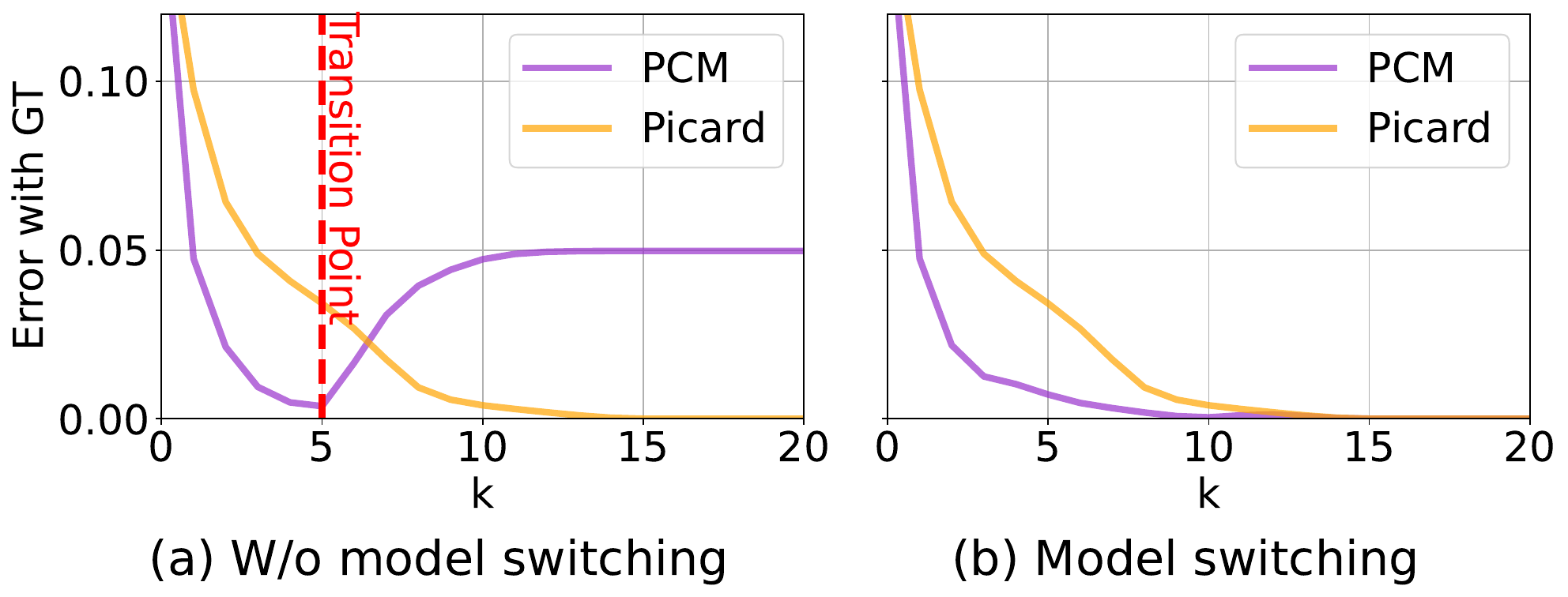}
    \caption{Comparison of convergence errors between Picard iteration at $k$ and the ground truth. (a) PCM w/o model switching converges faster than the naive Picard, but the error begins to increase at the \textit{transition point} and converges to a different point due to the modified weights. (b) PCM with model switching safely converges to the exact same output while ensuring accelerated convergence speed. Experiments are conducted on CelebA using DDIM.}
    \label{fig:nomix_versus_mix}
\end{figure}

However, because the PCT modifies the original weights, the convergent result of the \( \theta_{\text{PCM}} \) differs from the original model \( \theta_{\text{base}} \), sacrificing the primary advantage of parallel sampling. In Fig. \ref{fig:nomix_versus_mix} (a), we depict the error between the result of  picard iteration generated by \( \theta_{\text{PCM}} \) (purple) and the ground-truth solution generated by \( \theta_{\text{base}} \) (yellow). As shown in the figure, PCM initially converges to a lower error region faster than the baseline picard; however, it reaches a \textbf{\textit{transition point}} where the error increases, and the final convergence error remains greater than zero.

To address this issue, we propose the \textbf{Model Switching} method, which uses \( \theta_{\text{PCM}} \) in the early stages of the Picard iteration and switches to \( \theta_{\text{base}} \) in the later stages. Specifically, we employ a linear interpolation between the output estimated by \( \theta_{\text{PCM}} \) and \( \theta_{\text{base}} \), controlled by a mixing schedule \( \lambda(k) \), in our new Picard iteration, $X^{k+1}$ is computed as:

\begin{equation*}
    X^{k+1} \leftarrow  \lambda(k)*\Phi(X^k,\theta_{PCM}) + (1-\lambda(k))*\Phi(X^k,\theta_{base})
\end{equation*}

\begin{equation}
    \lambda(k) = \max(0,\min(1,1-s*k/K)). \label{eq:13}
\end{equation}

While we can use any decreasing function from 1 to 0 for \( \lambda(k) \), we employ the linear function with stiffness \( s \), controlling the switching speed of the schedule. $K$ denotes the predefined maximum iteration of picard iteration. As shown in Fig. \ref{fig:nomix_versus_mix} (b), with the Model Switching method, we can observe that the final convergence error approaches zero while maintaining the accelerated convergence speed.

\subsection{Training Stabilization with EMA}

\begin{figure}[t]
    \centering
        \begin{subfigure}[b]{0.45\columnwidth}
                \includegraphics[width=\textwidth]{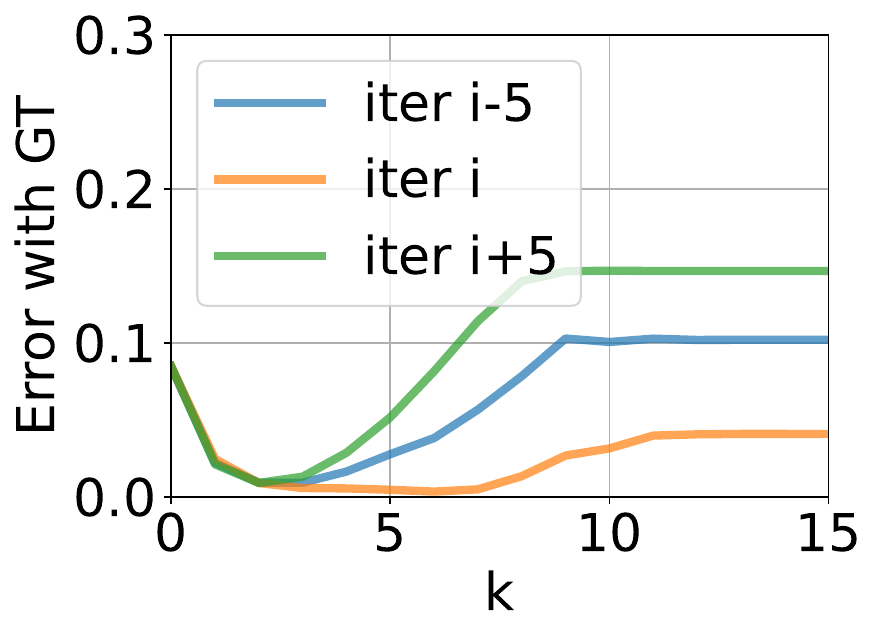}
                \caption{w/o EMA}
    \end{subfigure}
        \begin{subfigure}[b]{0.45\columnwidth}
                \includegraphics[width=\textwidth]{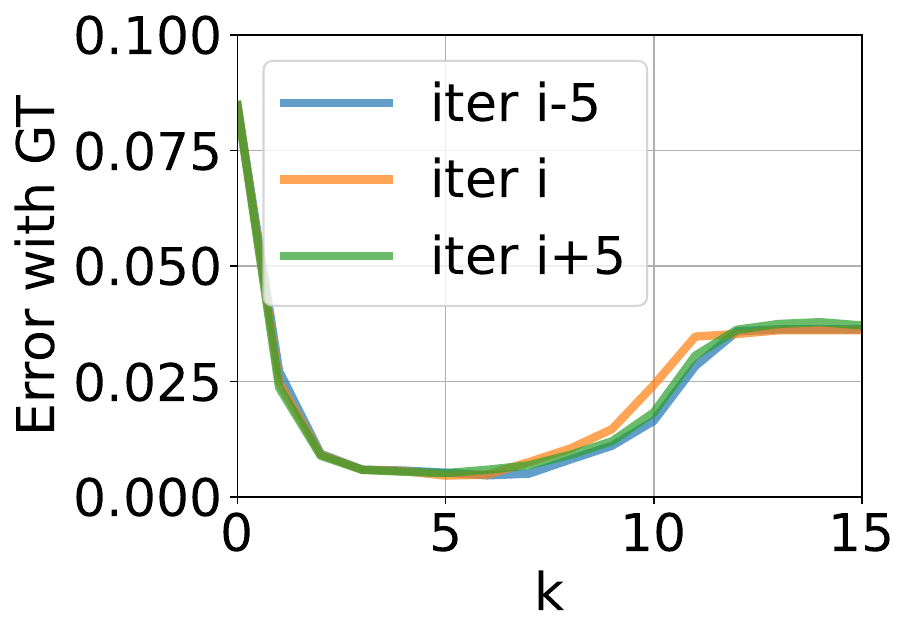}
                \caption{EMA}
    \end{subfigure}
    \caption{Training dynamics of PCM. (a) Naive PCT shows lots of training instability where transition point became different and final convergence error also different at every training iteration. (b) By using EMA, the error curve smoothly moved across the training iteration. Experiments are conducted on CelebA using DDIM.}
    \label{fig:dynamics_pcm}
\end{figure}

We found that PCT often destabilizes training, particularly affecting the transition point and final convergence of the Picard iteration. As shown in Fig. \ref{fig:dynamics_pcm} (a), the transition point and final convergence error of Picard iterations vary significantly with each training iteration. This variability makes it challenging to select the best model using the validation set, and it is especially difficult to predefine the model-switching schedule $\lambda(k)$, as the transition point changes with each training iteration. To address this, we follow prior research on stabilizing diffusion model training \cite{posthocema} and apply Exponential Moving Average (EMA) updates at each training iteration $i$, as shown below.

\begin{equation}
    \hat{\theta}_{\text{PCM}} \leftarrow \mu \hat{\theta}_{\text{PCM}} + (1 - \mu) \theta_{\text{PCM}}^i, \label{eq:14}
\end{equation}
where $\mu$ is a decay factor close to 1 (e.g., 0.999).
As shown in Fig \ref{fig:dynamics_pcm} (b), with EMA, we can notice that the transition point of $\theta_{PCM}$ and final convergence error moves smoothly across the training iteration.

\subsection{Parameter-space Model Switching with LoRA}
The remaining issues of PCT are that (1) it requires twice the model storage because both trained and untrained versions of the weights are needed, (2) the inference cost is doubled, even if parallelizable, and (3) full backpropagation of the diffusion model is required, which can be computationally intensive. To address these issues, we present an efficient version of our algorithm inspired by weight mixing \cite{modelsoup} and LoRA \cite{hu2021lora}. 

First, before training, we apply LoRA to every layer in the diffusion model. For example, for a linear layer $h = W_0 x$ where $W_0 \in \mathbb{R}^{d \times s}$, the LoRA $\Delta W$ is injected as
\begin{equation*}
    h = W_0 x + \Delta W x = W_0 x + BA x,
\end{equation*}
where $A \in \mathbb{R}^{r \times s}$ and $B \in \mathbb{R}^{d \times r}$ are the trainable up- and down-projection layers of LoRA, while $W_0$ freezed during training. Since typically $r \ll s,d$, the size of $\Delta W$ is usually much smaller than $\theta_{\text{PCM}}$.

During inference, now we apply Model Switching in the weight space rather than in the output space. Inspired by LoRA's weight-mixing strategy, we implement model switching by adjusting the scale $\lambda$ of LoRA weights according to picard iteration $k$.

\begin{equation}
    h^k = W_0 x^k + \lambda(k) \Delta W x^k = (W_0 + \lambda(k) \Delta W) x^k,
\end{equation}
where $h^k$ denotes the featue map $h$ at the $k$-th picard iteration
As with score mixing in the output space, $\lambda(k)$ is implemented using a linear schedule with stfiness $s$. This weight mixing operates quickly offline before each Picard iteration, allowing it to be implemented with the same latency as the original Picard iteration. 

In the experiment section, we assume that PCM uses feature-space model switching by default. We explicitly denote PCM-LoRA when LoRA-based parameter-space model switching is applied in addition to PCM.

\begin{algorithm}
  \caption{Picard Consistency Training}\label{algo:1}
  \begin{algorithmic}[1]
  \Require Original diffusion model $\theta_{base}$, ODE-Solver $\Phi$, picard step $K$, dataset size $N$
  \State $\triangleright\text{\ Dataset Generation}$
    \While{$n<N$}
        \State $\mathcal{J},k \gets \text{\{\}},0$ \Comment{Initialize} 
        \State $X^0\sim\mathcal{N}(0,I)$ \Comment{Sample initial noise} 
        \While{$k<K$}
            \State $X^{k+1} = \Phi(X^k)$ \Comment{Picard iteration} 
            \State $\text{Append\ } X^k \text{\ to\ } \mathcal{J}$, $k\gets k+1$
        \EndWhile
        \State Append $\mathcal{J}$ to $\mathcal{D}$, $n\gets n+1$
    \EndWhile\\ 

    \State $\triangleright\text{\ Training}$
    \State $\theta_{PCM},\hat{\theta}_{\text{PCM}} \gets \theta_{base}$ \Comment{Start from pretrained.} 
    \While {$\text{not converged}$}
        \State $\text{Sample\ }\mathcal{J}\sim\mathcal{D}, k\sim U[0,K-1]$
        \State $\text{Get\ }X^k,X^*\text{\ from\ } \mathcal{J}\text{and compute loss\ }\mathcal{L}\text{\ using Eq. \ref{eq:11} }$
        \State $\text{Backprop.\ }\theta_{PCM}\text{\ using\ }\mathcal{L}$
        \State $\text{Update\ }\text{EMA\ }\hat{\theta}_{\text{PCM}}\text{\ using Eq. \ref{eq:14} }$
    \EndWhile

       \State \Return $\hat{\theta}_{\text{PCM}}$
  \end{algorithmic}
\end{algorithm}

\begin{algorithm}
  \caption{Inference of Picard Consistency Model}\label{algo:2}
  \begin{algorithmic}[1]
  \Require Original diffusion model $\theta_{base}$, PCM $\theta_{PCM}$, ODE-Solver $\Phi$, picard step $K$
      \State $k \gets 0$ 
      \State $X^0 \sim N(0,I)$ \Comment{Sample initial noise}
       \While{$k < K$}
       \State $\lambda \gets \lambda(k)$ \Comment{Eq. \ref{eq:13}}
       \If{$\theta_{PCM}$\texttt{.use-lora}}
                  \For{\texttt{lora-layers\ }\textbf{in\ }$ \theta_{PCM}$}
        \State \texttt{lora-layers.scale $\gets\ \lambda$}
      \EndFor
      \State $X^{k+1} = \Phi(X^k;\theta_{PCM})$
        \Else
        \State $X_{PCM}, \gets \Phi(X^k;\theta_{PCM})$
        \State $X_{base} \gets \Phi(X^k;\theta_{base})$
        \State $X^{k+1} \gets \lambda*X_{PCM} + (1-\lambda)*X_{base}$
        \EndIf
        \State $k \gets k+1$
       \EndWhile
       \State \Return $X^K_T$ \Comment{Return last image}
  \end{algorithmic}
\end{algorithm}
\section{Experiments}
\label{sec:experiment}

To validate the effectiveness of our method, we conducted experiments across various tasks, including image generation and robot control. For image generation, we tested on the Latent Diffusion Model \cite{ldm} (LDM) , which uses a U-Net \cite{unet} architecture that operates denoising procedure in latent space of VAE, and performed experiments on the CelebA \cite{celeba} dataset and Stable Diffusion \cite{ldm} (SD), to conduct experiment on both unconditional and conditional sampling cases. For robotic control, we follow setups in Diffusion Policy \cite{diffusionpolicy}. Detailed experimental setups are provided in the relevant section.

For evaluation, we performed qualitative and quantitative analysis in both tasks. For the qualitative experiments, in image generation, we demonstrate that our method achieves generated exactly the same results compared to sequential denoising and our PCM generates higher-quality images within the same number of iterations compared to the naive Picard \cite{shih2024parallel}. In robotic control tasks, we visualize an episode of pushT tasks using naive Picard and our PCM, demonstrating PCM safely generate correct motion sequences while naive Picard failed. 

In the quantitative experiments, for CelebA, we measured the latency speedup and Fréchet Inception Distance (FID) \cite{heusel2017gans} using 5,000 generated images. For stable diffusion, we report the theoretical speedup and CLIP score \cite{hessel2021clipscore} to measure the correctness of image and given prompt. For robotic control tasks, we report the latency speedup and average reward of episodes.
The Picard Consistency Training and dataset generation are conducted on high-performance servers equipped with 8×RTX3090 and evaluated on 1×A6000 GPUs. We use CUDA 11.8 and Pytorch 2.4.0 for experiments.

\begin{figure}[t!]
    \centering
        \begin{subfigure}[b]{0.32\columnwidth}
                \includegraphics[width=\textwidth]{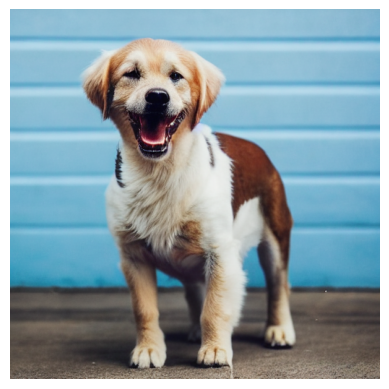}
                \caption{Sequential}
    \end{subfigure}
        \begin{subfigure}[b]{0.32\columnwidth}
                \includegraphics[width=\textwidth]{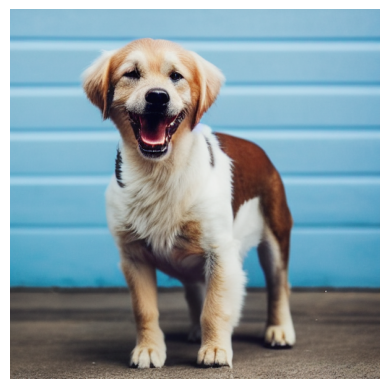}
                \caption{Picard}
    \end{subfigure}
        \begin{subfigure}[b]{0.32\columnwidth}
                \includegraphics[width=\textwidth]{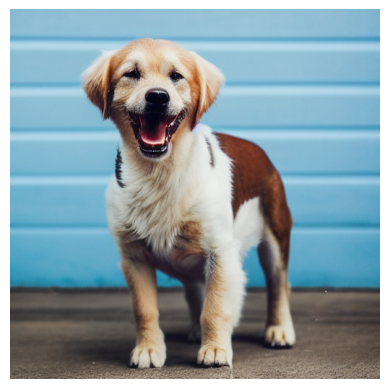}
                \caption{PCM}
    \end{subfigure}
    \caption{Qualitative comparison of Sequential (a), Picard (b), and PCM (c) on Stable Diffusion v1.4 using DDIM. All methods converge to the exact same solution, with PCM requiring 2.8 times less latency than Sequential.}
    \label{fig:converg_imgen}
\end{figure}

\begin{figure}[t!]
    \centering
        \begin{subfigure}[b]{0.98\columnwidth}
                \includegraphics[width=\textwidth]{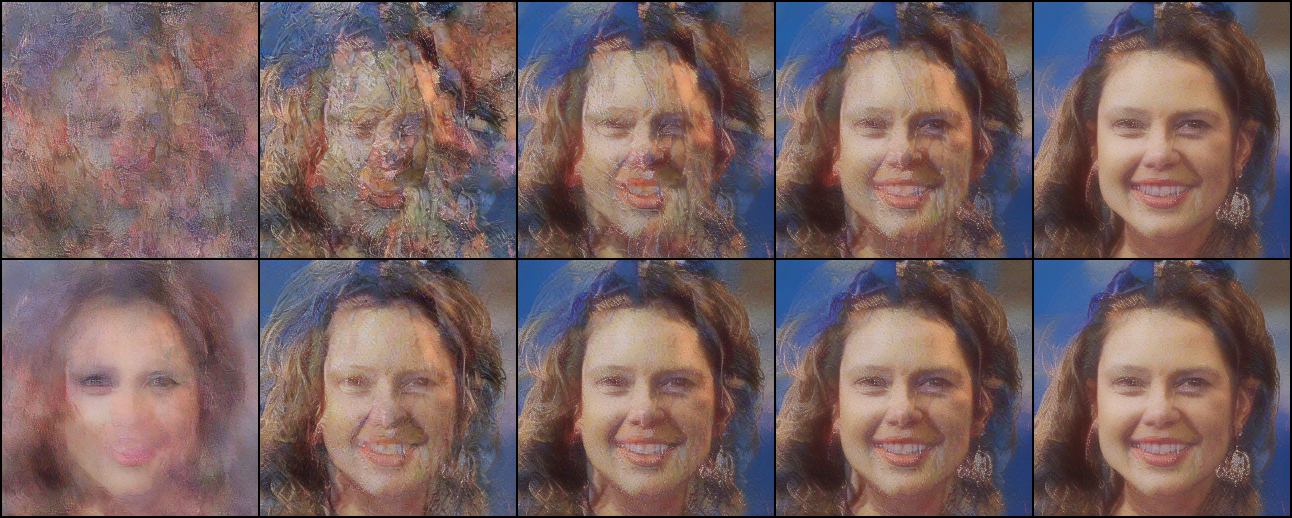}
    \end{subfigure}
    \caption{Qualtitive comparison of Picard (up-row) and PCM(down-row) in same iteration $k$ on LDM-CelebA with DDIM. While Picard generate inaccurate images in early iteration,  PCM generate data that are more closely resembles actual images.
    }
    \label{fig:qual_imgen}
\end{figure}

\subsection{Image Generation}
First, we conducted quantitative experiments on image generation task by measuring FID, CLIP score, and latency speedup. For CelebA, we generated 500 samples for the dataset for Picard Consistency Training and trained the model over 50 epochs using the Adam optimizer with a learning rate of 1e-4. The final model was selected based on the highest performance across 10 noise configurations used for evaluation. We use total time step $T$ for 50 for picard iteration. For SD, we setup similar experimental setup and report the theoretical speed-up using ratio of sequential steps. We depicts details and hyperparameters in our Appendix. For PCM-LoRA, following prior work \cite{hu2021lora}, we applied LoRA to the linear layers in attention in U-Net.

As shown in Table \ref{tab:performance_ldm}, for LDM-CelebA, while sequential DDIM requires 18 steps to converge, PCM converges in just 6 steps, achieving a speedup of 2.11x. Although the Picard \cite{shih2024parallel} also converges in a similar number of steps, PCM consistently achieves a lower FID than Picard. Furthermore, as the sequential step \( k \) in Picard iterations decreases, Picard \cite{shih2024parallel} generates lower-quality images, whereas PCM rapidly generates feasible image quality. For example, while Picard produces entirely incorrect images at 2 steps with an FID of 257.82, PCM generates images with FID of 67.74.

Table \ref{tab:performance_sd} presents similar results for SD. As shown, PCM consistently produces better CLIP score than Picard under the same sequential iteration \( k \), even surpassing the performance of Picard at \( k = 9 \) when using PCM-LoRA at \( k = 8 \).

We also depict the qualitative comparison of Picard and PCM. As shown in Fig. \ref{fig:converg_imgen}, both Picard (Fig. \ref{fig:converg_imgen} (b)) and PCM (Fig. \ref{fig:converg_imgen} (c)) converged to the exact same image as the ground truth of sequential denoising in Fig. \ref{fig:converg_imgen} (a). This result demonstrates that our PCM successfully restores the output distribution of the original model through model switching, even when starting with different weights. In Fig.~\ref{fig:qual_imgen}, we present the results of the naive Picard (up-row) and our PCM (down-row) for the same sequential Picard iteration \( k \). As seen in the figure, especially at low \( k \), Picard generates almost incorrect images, while PCM produces data that more closely resembles actual images.

\begin{table}[t!]
\centering
\resizebox{0.92\columnwidth}{!}{%
\begin{tabular}{c|c|c|c|c}
\hline
\multicolumn{5}{c}{\textbf{LDM-CelebA}} \\
\midrule
\hline
(DDIM) & \textbf{Steps}$\downarrow$ & \textbf{FID}$\downarrow$ & \textbf{latency}$\downarrow$ & \textbf{Speedup}$\uparrow$ \\
\hline
Sequential\cite{ddim} &  9 & 51.97 & 1.42s & 1x \\
Picard\cite{shih2024parallel} & 4 & 60.82& 0.88s & 1.61x \\
\textbf{PCM} & 3 & 56.80 & 0.91s & 1.56x \\
\textbf{PCM-LoRA} & 3 & \textbf{50.10} & \textbf{0.66s} & \textbf{2.15x} \\
\hline
Sequential\cite{ddim} & 15 & 38.87 & 2.36s & 1x \\
Picard\cite{shih2024parallel} & 5 & 42.67 & 1.10s & 2.14x \\
\textbf{PCM} & 5 & 40.09 & 1.52s & 1.55x \\
\textbf{PCM-LoRA} & 5 & \textbf{38.35} & \textbf{1.10s} & \textbf{2.14x} \\
\hline
Sequential\cite{ddim} & 18 & 36.09 & 2.83s & 1x \\
Picard\cite{shih2024parallel} & 6 & 36.19 & 1.34s & 2.11x\\
\textbf{PCM} & 6& 36.67 & 1.87s & 1.51x\\
\textbf{PCM-LoRA} & 6 & \textbf{35.97} & 1.34s & 2.11x \\
\hline
\multicolumn{5}{c}{Comparison under same sequential step} \\
\hline
Sequential\cite{ddim} & 2 & 366.92 & 0.32s & - \\
Picard\cite{shih2024parallel} & 2 & 257.83 & 0.44s & -\\
\textbf{PCM} & 2 & 77.62 & 0.60s & -\\
\textbf{PCM-LoRA} & 2 & \textbf{67.74} & 0.44s & -\\
\midrule
\hline
(DPM-Solver) & \textbf{Steps}$\downarrow$ & \textbf{FID}$\downarrow$ & \textbf{latency}$\downarrow$ & \textbf{Speedup}$\uparrow$ \\
\hline
Sequential & 7 & 64.06 & 1.21s & 1x \\
Picard & 5 & 64.68 & 1.10s & 1.09x\\
\textbf{PCM} & 4 & 64.14 & 1.20s & 1.01x \\
\textbf{PCM-LoRA} & 4  & \textbf{62.61} & \textbf{0.88s} & \textbf{1.37x}\\
\hline
\end{tabular}%
}
\caption{Image generation performance comparison on LDM-CelebA dataset.}
\label{tab:performance_ldm}
\end{table}

\begin{table}[h!]
\centering
\resizebox{\columnwidth}{!}{%
\begin{tabular}{c|c|c|c}
\hline
\multicolumn{4}{c}{\textbf{Stable Diffusion(SD) V1.4}} \\
\midrule
\hline
(DDIM) & \textbf{Steps}$\downarrow$ & \textbf{CLIP-Score}$\uparrow$ &  \textbf{Theoretical Speedup}$\uparrow$ \\
\hline
Sequential\cite{ddim} & 14 & 26.20 & 1x \\
Picard\cite{shih2024parallel} & 5 & 26.24 & 2.8x \\
\textbf{PCM} & 5 & 26.25  & 2.8x \\
\textbf{PCM-LoRA} & 5 & \textbf{26.26} & \textbf{2.8x} \\
\hline
Sequential\cite{ddim} & 25& 26.34 & 1x\\
Picard\cite{shih2024parallel} & 9 & 26.34 & 2.78x \\
\textbf{PCM} & 9 & 26.34  & 2.78x \\
\textbf{PCM-LoRA} & 8 & \textbf{26.35}  & \textbf{3.13x} \\
\hline
\multicolumn{4}{c}{Comparison under same sequential step} \\
\midrule
Sequential\cite{ddim} & 4 & 25.00 & -\\
Picard\cite{shih2024parallel} & 4 & 25.91  & - \\
\textbf{PCM} & 4 & 25.92  & - \\
\textbf{PCM-LoRA} & 4 & \textbf{26.02}  & - \\
\hline
\end{tabular}%
}
\caption{Image generation performance comparison on Stable Diffusion v1.4.}
\label{tab:performance_sd}
\end{table}

\begin{table}[t]
\centering
\resizebox{0.92\columnwidth}{!}{%
\begin{tabular}{c|c|c|c|c}
\hline
\multicolumn{5}{c}{\textbf{PushT}} \\
\midrule
\hline
(DDIM) & \textbf{Steps}$\downarrow$ & \textbf{Reward}$\uparrow$ & \textbf{Time }$\downarrow$ & \textbf{Speedup}$\uparrow$\\
\hline
Sequential\cite{diffusionpolicy} & 25 & 0.80 & 5.14s & 1x \\
Picard\cite{shih2024parallel} & 15 & 0.80 & 3.57s  & 1.44x \\
\textbf{PCM} & \textbf{8} & \textbf{0.83} & 2.01s & 2.55x  \\
\textbf{PCM-LoRA} & \textbf{8} & \textbf{0.83}  & 1.90s & \textbf{2.71x} \\
\hline
Sequential\cite{diffusionpolicy} & 7 & 0.45 & 1.44s & 1x \\
Picard\cite{shih2024parallel} & 5 & 0.43 & 1.19s  & 1.21x \\
\textbf{PCM} & \textbf{5} & \textbf{0.48} & 1.25s & 1.15x \\
\textbf{PCM-LoRA} & \textbf{5} & \textbf{0.47} & 1.19s & \textbf{1.21x}  \\
\hline
Sequential\cite{diffusionpolicy} & 4 & 0.12 & 0.82s & 1x \\
Picard\cite{shih2024parallel} & 3 & 0.12 & 0.71s & 1.15x \\
\textbf{PCM} & \textbf{3} & \textbf{0.15} & 0.75s & \textbf{1.10x}  \\
\textbf{PCM-LoRA} & \textbf{3} & \textbf{0.15} & 0.71s & \textbf{1.15x}  \\
\hline
\end{tabular}%
}
\caption{Robotic control performance comparison on PushT dataset.}
\label{tab:performance_pusht}
\end{table}
\subsection{Robotic Control}
Recently, research on creating robust robot action sequences using diffusion models has been showing increasingly promising results \cite{diffusionpolicy}. Compared to traditional methods, diffusion models demonstrate high robustness and quality in motion generation across various situation and modalities. However, since increasing the speed of motion control is crucial for improving industrial productivity and user satisfaction, the high latency from the iterative nature of diffusion models poses a significant obstacle to their commercialization. Therefore, we experimented our PCM with a latency-critical robotic control task. Following setups from Diffusion Policy \cite{diffusionpolicy}, a recently proposed diffusion model framework for behavioral cloning, we conduct an experiment using the PushT dataset, which aims to generate motion sequences for locating the T-block in the desired location. We used 1D state based U-Net that takes observation of current state as input and use action horizon with 8 and maximum trajectory length with 200 for the experiment.
For PCT, we generated a total of 2000 samples for datasets and use Adam optimizer with learning rate of 1e-4 for 10 epochs.

\begin{figure}[t]
    \centering
        \begin{subfigure}[b]{0.98\columnwidth}
                \includegraphics[width=\textwidth]{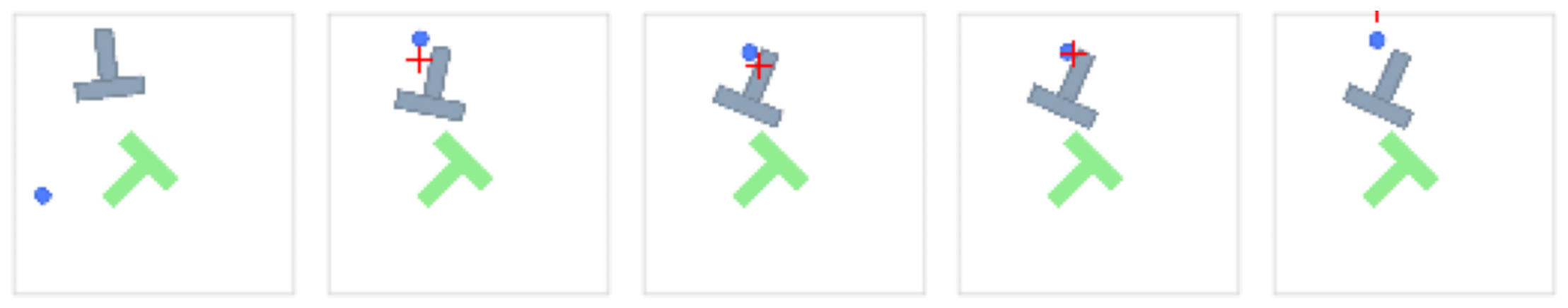}
                \caption{Picard}
    \end{subfigure}
        \begin{subfigure}[b]{0.98\columnwidth}
                \includegraphics[width=\textwidth]{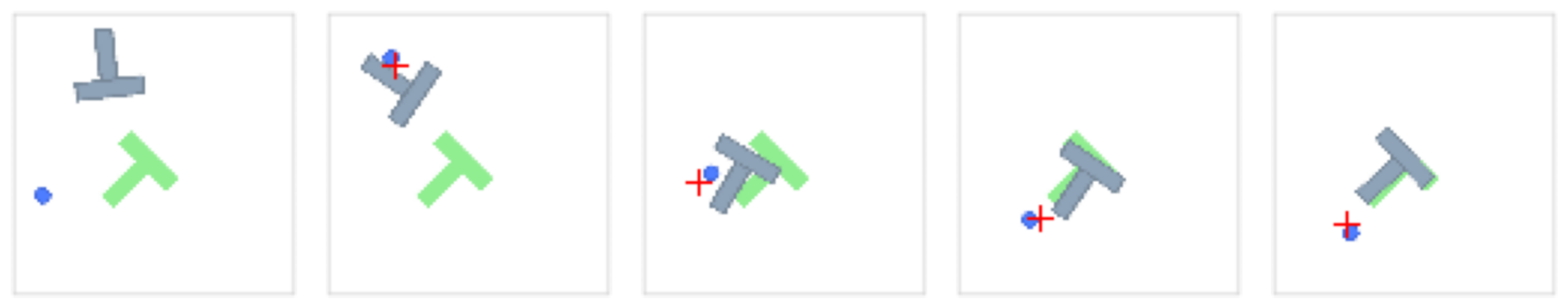}
                \caption{PCM}
    \end{subfigure}

    \caption{Qualitative comparison of Picard and PCM at PushT dataset. We use DDIM sampler and record 1 episode using $k$=6.  (a) While naive Picard fails to locate the T-block in correct location (b) PCM successfully completes the task.}
    \label{fig:qual_pusht}
    \vspace{-0.3cm}
\end{figure}

In Table \ref{tab:performance_pusht}, we present the average reward and latency speedup of sequential, Picard, and PCM methods using DDIM on the PushT task. As shown in the table, while the Sequential method requires 25 steps to reach a reward of 0.8, PCM converges in only 8 steps, achieving a even higher reward of 0.83 with 2.7x wall-clock speed-up and approximately twice the acceleration compared to Picard iteration. Even at lower reward levels, PCM achieves a higher reward with faster speed-up than Sequential and surpasses Picard iteration with the same number of steps. 
We also depict the qualitative comparison of Picard and PCM in Fig. \ref{fig:qual_pusht}. As shown in the figure, under the same latency budget (the same k), (a) While naive Picard fails to locate T-block in correct location, (b) PCM successfully completes the task.

\subsection{Ablation Study}

In this section, we conducted an ablation study to analyze the impact of different components in our PCM. Specifically, we examined the effects of loss weighting and model switching on the feature and weight spaces with different values of stiffness \( s \).

\subsubsection{Effect of Loss Weighting}

In Table~\ref{table:alpha_effect}, we compare the results with and without the loss-weighting term  \( \alpha(k) \)  from Eq.~\ref{eq:11}. As shown in the table, without loss weighting, the model converges to a suboptimal solution due to the variance gap in the Picard iterations. This result highlights the importance of applying loss weighting based on variance levels, similar to the approach used in standard diffusion model training.

\begin{figure}[t!]
    \centering
        \includegraphics[width=0.99\linewidth]{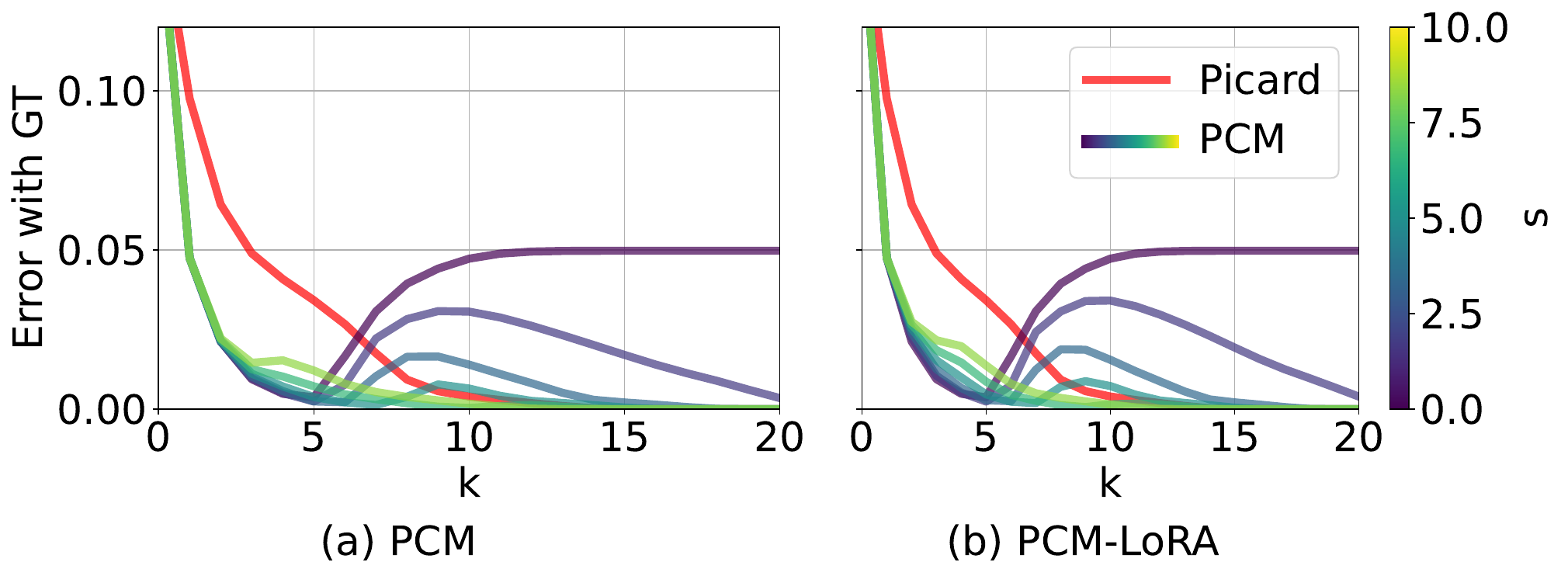}

        \caption{Effect of model switching on different stiffness $s\in\{0,2,4,6,8,10\}$. Experiments are conducted on LDM-CelebA with DDIM sampler.}
            \label{fig:model_switching}
\end{figure}

\subsubsection{Effect of Model Switching} 
In Fig.~\ref{fig:model_switching}, we compare the convergence error between the ground truth and PCM iterations at various \( k \) values, using different stiffness \( s \) from Eq.~\ref{eq:13}. As shown, both (a) the feature space and (b) the weight space converge to the zero-error region when an appropriate \( s \) is used. However, when \( s \) is too small, the error converges to a non-zero point, and when \( s \) is too large, the error curve fluctuates slightly between \( k = 3 \) and \( k = 4 \), resulting in slower convergence.

\subsection{Comparision with Newton's Method}
In this section, we conducted a comparison experiment with Newton's method \cite{polyak2007newton}, a well-known technique for accelerating the root-finding problem. Specifically, finding the solution to the fixed-point problem can be converted into a root-finding problem: \( \Psi(X^*) = \Phi(X^*) - X^* = 0 \). To solve this root-finding problem, we apply Newton's method: \( X^{k+1} \leftarrow X^k - J_{X^k}[\Psi(X^k)]^{-1} \Psi(X^k) \), where \( J[\cdot] \) is the Jacobian. While computing the Jacobian inverse is generally infeasible for large-scale problems like neural networks, we can emulate Newton's method with \( O(T) \) sequential time by leveraging the MCMC properties of the denoising process. Although this approximated method does not improve latency, it allows us to compare the speed of convergence and assess the potential margin for PCM. Details of the Newton algorithm are provided in the Appendix.

In Fig. \ref{fig:newton_fig}, we compare the convergence results of Newton's method and PCM on the LDM-CelebA dataset. As shown, the points where Newton's method converges at a low error level are nearly the same as the convergence points of PCM, indicating that the convergence speed of PCM is actually close to that of the oracle. Additionally, unlike Newton's method, which often struggles with poor initialization, PCM  shows stable convergence.

\begin{figure}[t!]
    \vspace{-0.4cm}
    \centering
    \begin{minipage}{0.42\linewidth}
        \begin{table}[H]
            \centering
            \begin{tabular}{c|c}
                \toprule
                \midrule
                $\alpha(k)$ & Reward $\uparrow$ \\
                \hline
                \midrule
                1 & 0.71 \\ 
                \hline
                $1/\sqrt{var(k)}$ & \textbf{0.83} \\
                \hline
                $1/k$ & 0.78 \\
                \hline
                \midrule
            \end{tabular}
            \caption{Effect of $\alpha(k)$. Experiments are conducted with PushT task on $k=8$.}
            \label{table:alpha_effect}
        \end{table}
    \end{minipage}%
    \hfill
    \begin{minipage}{0.52\linewidth}
        \begin{figure}[H]
            \centering
            \includegraphics[width=\linewidth]{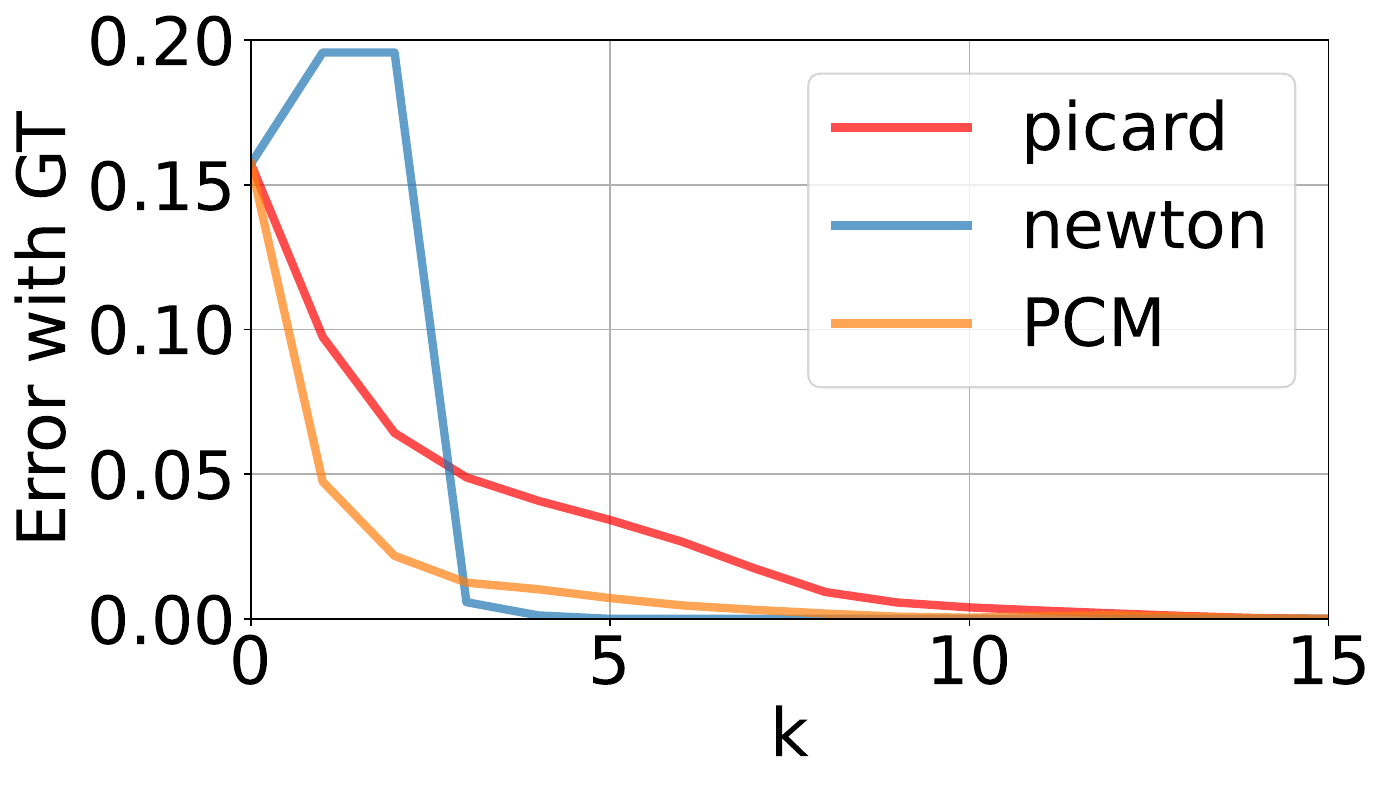}
            \caption{Comparison with Newton's method. We use LDM-CelebA and DDIM 50-step.}
            \label{fig:newton_fig}
        \end{figure}
    \end{minipage}
\end{figure}

\section{Related Works}
\label{sec:rel_works}

\subsection{Efficient Diffusion Models}
To accelerate the slow generation speed of diffusion models, various studies have introduced methods to create more efficient diffusion models. One of the most active research areas focuses on developing differential equation solvers \cite{ddim, dpmsolver, dpmpp, xue2024sa}, which aim to preserve generation quality while reducing the number of sampling steps. Additionally, to achieve extremely low sampling steps, distillation-based approaches \cite{song2023consistency, salimans2022progressive, guided} have also been widely studied. Alternatively, to reduce the computational cost, optimization methods such as quantization \cite{qdiffusion, huang2024tfmq} and caching techniques \cite{ma2024deepcache, so2023frdiff} are also being actively explored.

\subsection{Parallel inference of Sequential Models}

Several pioneering studies have aimed to reduce the latency of sequential models by leveraging parallel computation. An early study \cite{song2021accelerating} introduced a parallel inference algorithm for sequential models using Jacobi and Gauss-Seidel iterations. Building on this foundation, \cite{lim2023parallelizing} proposed an accelerated parallel inference method for Jacobi iteration using quasi-newton methods. The ParaDiGMS framework \cite{shih2024parallel} was the first to apply parallelization to denoising steps in diffusion models through Picard iteration. Furthermore, ParaTAA \cite{tang2024accelerating} reframed fixed-point Picard iterations as a nonlinear root-finding problem, proposing an acceleration technique based on Anderson Acceleration. However, Picard iteration often suffers from slow convergence in practice, and ParaTAA requires additional computation and memory for the quasi-newton matrix during inference. Parallel inference for language models has also become an active research area, with methods like speculative decoding \cite{leviathan2023fast, kim2024speculative, sun2024spectr, li2024eagle} and Jacobi decoding \cite{song2021accelerating, fu2024break}. Recently, similar to our approach, CLLM \cite{kou2024cllms} accelerated jacobi decoding by minimizing the distance between a random trajectory and the fixed-point solution.

\section{Conclusion}
\label{sec:conclusion}

In this paper, we introduce the Picard Consistency Model (PCM) to enhance parallel sampling in diffusion models. Inspired by Consistency Models, our Picard Consistency Training enables the diffusion model to predict the final fixed-point solution at each Picard iteration, accelerating convergence notably. Additionally, we propose a model-switching method to preserve the exact convergence properties of Picard iteration. Extensive experiments demonstrate that PCM achieves approximately a 2.71x speedup compared to sequential denoising and around a 1.77x speedup over standard Picard iteration without compromising generation quality.

\paragraph{Acknowledgments}   

This work was supported by IITP and NRF grants funded by the Korea government(MSIT) (Nos. RS-2019-II191906, RS-2023-00213611, RS-2024-00415602, RS-2024-00457882).

{
    \small
    \bibliographystyle{ieeenat_fullname}
    \bibliography{main}
}

 \clearpage
\setcounter{page}{1}
\maketitlesupplementary

\renewcommand{\thesection}{S\arabic{section}}
\renewcommand{\thefigure}{S\arabic{figure}}
\renewcommand{\thetable}{S\arabic{table}}

\renewcommand{\theequation}{S\arabic{equation}}

\setcounter{section}{0}
\setcounter{figure}{0}
\setcounter{table}{0}
\setcounter{equation}{0}

\section{Experimental Setup}
\label{sec:hyp}
In this section, we provide detailed hyperparameter configurations for Picard Consistency Training (PCT) on the CelebA, Stable Diffusion, and PushT datasets.

All experiments in this paper were conducted on GPU servers equipped with 8 NVIDIA GTX 3090 GPUs (each with 24 GB VRAM) and 2 AMD 7313 CPUs. The implementations were built using the PyTorch v2.4.0 and CUDA 11.8 framework. The source code for our work is included in the supplementary materials.

In Table \ref{tab:hyp_img}, we summarize the hyperparameter configurations for PCT in the image generation tasks. Additionally, we applied Low-Rank Adaptation (LoRA) to all attention layers in our 2D-UNet architecture. For the construction of the training dataset for Stable Diffusion, we selected 20 arbitrary prompts and extracted 10 images for each prompt. For the CLIP score evaluation, we used 200 images generated from the prompts used during training and 300 images generated from unused prompts.

Table \ref{tab:hyp_rob} also provides the hyperparameter configurations for PCT in the robotic control task, PushT. Moreover, the diffusion model for the robotic task takes the observation as input and predicts the action sequence over a defined action horizon. The model then executes the predicted action and uses the subsequent observation as input for the next step. We set the action horizon to 8, with a maximum of 200 steps in the action sequence generation. For this task, we use a 1D-state-based UNet and also apply LoRA to the attention layers within the transformer blocks.

\begin{table}[h!]
\centering
\resizebox{\columnwidth}{!}{%
\begin{tabular}{c|c|c|c}
\hline
\hline
\multicolumn{4}{c}{\textbf{CelebA}} \\
\hline
\hline
\textbf{\# dataset} & \textbf{T} & \textbf{K} & \textbf{Solver} \\
\hline
200 & 50 & 30 & DDIM,DPMSolver \\
\hline
\textbf{Optimizer} & \textbf{learning rate} & \textbf{scheduler} & \textbf{epoch} \\
\hline
Adam & 1e-4 & Cosine & 50 \\
\hline
\hline
\multicolumn{4}{c}{\textbf{Stable Diffusion v1.4}} \\
\hline
\hline
\textbf{\# dataset} & \textbf{T} & \textbf{K} & \textbf{Solver} \\
\hline
200 & 30 & 15 & DDIM \\
\hline
\textbf{Optimizer} & \textbf{learning rate} & \textbf{scheduler} & \textbf{epoch} \\
\hline
Adam & 1e-4 & Cosine & 10 \\
\hline
\end{tabular}%
}
\caption{Hyperparameters for PCT in image generation tasks.}
\label{tab:hyp_img}
\end{table}

\begin{table}[h!]
\centering
\resizebox{0.8\columnwidth}{!}{%
\begin{tabular}{c|c|c|c}
\hline
\hline
\multicolumn{4}{c}{\textbf{PushT}} \\
\hline
\hline
\textbf{\# dataset} & \textbf{T} & \textbf{K} & \textbf{Solver} \\
\hline
2000 & 50 & 30 & DDIM \\
\hline
\textbf{Optimizer} & \textbf{learning rate} & \textbf{scheduler} & \textbf{epoch} \\
\hline
Adam & 5e-5 & Cosine & 200 \\
\hline
\end{tabular}%
}
\caption{Hyperparameters for PCT in robotic control tasks.}
\label{tab:hyp_rob}
\end{table}

\section{Procedure of emulating Newton's method}
\label{sec:newton}
As mentioned in the paper, our objective is to find the solution to the fixed-point problem $\Phi(X^*) = X^*$, where $x \in \mathbb{R}^d$ and $X = [x_0, x_1, \dots, x_{T-1}] \in \mathbb{R}^{T*d}$. This fixed point problem can be reformulate as root-finding problem $\Psi(X^*) - X^* = 0$. The \textit{inference} of $\Phi,\Psi$ are defined as 
\begin{equation}
    \Phi(X)=[x_0, F(t_1,x_0), F(t_2,x_1), \dots, F(t_{T-1}, x_{T-2})]
\end{equation}
\begin{equation}
    \Psi(X)=[0, x_1 - F(t_1,x_0) \dots, x_{T-1} - F(t_{T-1}, x_{T-2})]
\end{equation}

Where $F(\cdot)$ is single inference of diffusion model.
We start by apply newton's method to solve root-findinng problem $\Psi(\cdot)$: 

\begin{equation}
    X^{k+1} = X^k - J_{X^k}[\Psi(X^k)]^{-1} \Psi(X^k),
\end{equation}

where $J_{X^k}[\Psi(X^k)]$ denotes the Jacobian matrix of $\Psi(X^k)$ with respect to $X^k$.

However, directly computing the Jacobian matrix and its inverse becomes computationally prohibitive for large-scale problems, such as diffusion models. For instance, in the case of the CelebA dataset, $d = 32 \times 32 \times 3 = 3072$ and $T = 50$, resulting in a Jacobian matrix of size $(d \cdot T) \times (d \cdot T) = 153600 \times 153600$, requiring approximately 700 GB of memory to store the Jacobian matrix.

To implicitly emulate the Newton iteration, we start by multiplying the Jacobian matrix on both sides:

\begin{equation}
    J_{X^k}[\Psi(X^k)] X^{k+1} = J_{X^k}[\Psi(X^k)] X^k - \Psi(X^k).
\end{equation}

Rearranging terms, we get:

\begin{equation}
    J_{X^k}[\Psi(X^k)] (X^{k+1} - X^k) = -\Psi(X^k).
\end{equation}

In matrix form, this becomes:

\begin{equation}
\begin{array}{c}
\begin{bmatrix}
    \frac{\partial F(t_1, x_0^k)}{\partial x_0^k} & I & 0 & \dots & 0 \\
    0 & \frac{\partial F(t_2, x_1^k)}{\partial x_1^k} & I & \dots & 0 \\
    \vdots & \vdots & \ddots & \ddots & \vdots \\
    0 & 0 & \dots & \dots & I
\end{bmatrix}
\cdot
\begin{bmatrix}
    x_1^{k+1} - x_1^k \\
    x_2^{k+1} - x_2^k \\
    \vdots \\
    x_{T}^{k+1} - x_{T}^k
\end{bmatrix}
\\
=
\begin{bmatrix}
    F(t_1,x_0^k) - x^k_1 \\
    F(t_2,x_1^k) - x^k_2  \\
    \vdots \\
    F(t_{T},x_{T-1}^k) - x^k_{T} 
\end{bmatrix}
\end{array}
\end{equation}

solving this equation is equivalent as : 

\begin{equation}
    x^{k+1}_{t_{i+1}} = F({t_{i+1}}, x_t^k) + \frac{\partial F({t_{i+1}}, x_{t_i}^k)}{\partial x_{t_i}^k}(x_{t_i}^{k+1} - x_{t_i}^k)
\end{equation}

As shown in the equation, the computation of $\frac{\partial F(t_{i+1}, x_{t_i}^k)}{\partial x_{t_i}^k} (x_{t_i}^{k+1} - x_{t_i}^k)$ corresponds to a Jacobian-vector product (JVP), which can be efficiently computed using autograd frameworks such as PyTorch. By applying this equation and computing the JVP sequentially for each $t_i$, we can emulate single iteration of Newton's method in $O(T)$ sequential time with feasible computational resources.

\begin{figure}[h!]
    \centering
    \includegraphics[width=1\linewidth]{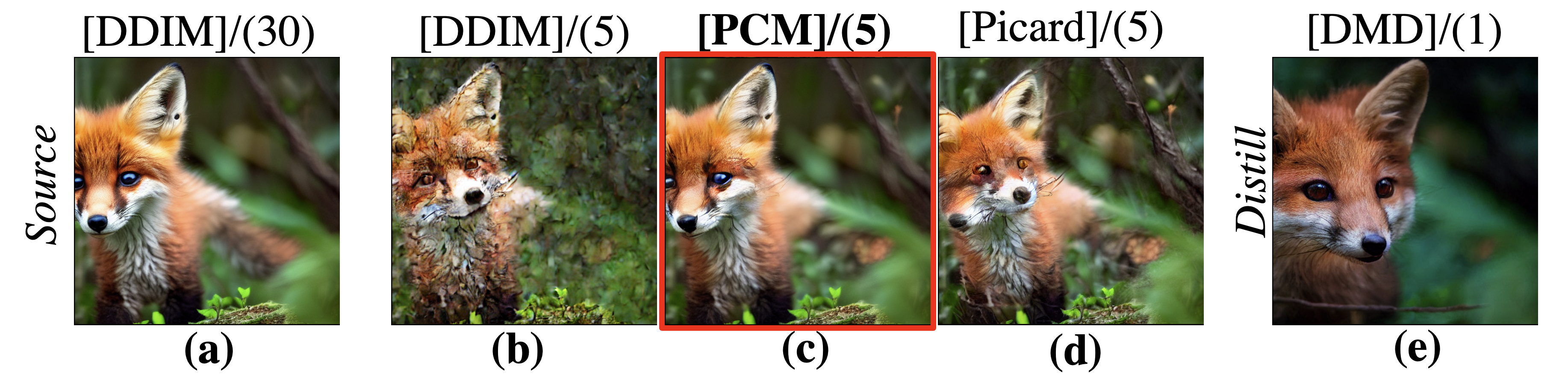}
    \caption{ [x]/(y) denotes [Method]/(Sequential Step)}
    \label{fig:enter-label}
\end{figure}

\section{Comparison with Distillation-based methods}
\label{sec:distill}
We avoid direct comparison between PCM with distillation-based methods(\textit{Distill}), such as CM, DMD, and CTM, since the two approaches have different goals. \textit{Distill} sacrifices diversity (and quality in some cases) for lower latency, which generates different output from the source model due to modified weights. See Fig.\ref{fig:enter-label}(a)(source) and (e)(\textit{Distill}). This problem can be crucial in scenario where even minor quality degradation is unacceptable, such as robotic control. 
In contrast, PCM accelerates inference \textbf{without any quality degradation}, as shown in Fig.\ref{fig:enter-label} (c). This exact convergence property is guaranteed by the Picard theorem, making PCM apart from typical optimization methods that trade latency for quality. In fact, many parallel sampling works \cite{shih2024parallel,tang2024accelerating} did not compare against \textit{Distill} for this reason. Furthermore, PCM offers advantages in training cost: \textit{Distill} typically requires days, whereas PCM takes only an hour.

\section{Effect of weight-mixing}
\label{sec:wmeffect}

In Table~\ref{tab:effectwm}, we present the effects of weight mixing to better highlight the novelty of PCM. As shown, without weight mixing, the FID improvement of PCM over Picard is marginal, even resulting in a higher FID score than the naive Picard. However, when weight mixing is applied, PCM achieves a significantly better FID score, outperforming Picard.

\begin{table}[h]
    \centering
    \renewcommand{\arraystretch}{0.99} 
    \setlength{\tabcolsep}{1pt} 
\footnotesize
    \begin{tabular}{c|ccccc}
    \hline
        LDM-CelebA (FID$\downarrow$) &  k=1 & k=2 & k=3 & k=4 & k=5 \\
        \hline
        Picard & 382.48 & 257.83 & 109.67 & 60.82 & 42.66 \\
        \hline
        PCM w/o Mix & 124.93& 76.75 & 63.54 & 54.46 & 50.07 \\
        PCM w. Mix & \textbf{124.93} & \textbf{67.77} & \textbf{50.14} & \textbf{41.94} & \textbf{38.36} \\
        \hline
    \end{tabular}
    \caption{Effect of Weight Mixing}
    \label{tab:effectwm}
\end{table}

\section{Evaluation on various metric}
\label{sec:metric}
In Table~\ref{tab:metric}, we report the results of multiple evaluation metrics for generative models—including FID, sFID, IS, Precision, Recall, and CMMD—measured on LDM-CelebA with $k=5$. The results demonstrate that PCM consistently outperforms both Picard and Sequential across all metrics, except for IS, which is often considered \textit{noisy} as it reflects the variance of the generated images.

\begin{table}[h]
    \centering
     \renewcommand{\arraystretch}{0.99} 
    \setlength{\tabcolsep}{1pt} 
    \footnotesize
    \begin{tabular}{c|cccccc}
    \hline
        Methods & FID$\downarrow$ & sFID$\downarrow$ & IS$\uparrow$ & Precision$\uparrow$ & Recall$\uparrow$ & CMMD$\downarrow$ \\
        \hline
        Sequential & 375.08 & 54.75 & 2.57 & 0.00 & 0.00 & 4.95 \\
        Picard & 42.66 & 20.36 & \textbf{3.07} & 0.43 & 0.37 & 1.95 \\
        PCM & \textbf{38.36} & \textbf{15.13} & 2.76 & \textbf{0.44} & \textbf{0.42} & \textbf{1.71} \\
        \hline
    \end{tabular}
    \caption{More metric evaluations}
    \label{tab:metric}
\end{table}

\section{Pareto Front Comparison}
\label{sec:pareto}

\begin{figure}[t]
    \centering
        \begin{subfigure}[b]{0.45\columnwidth}
                \includegraphics[width=\textwidth]{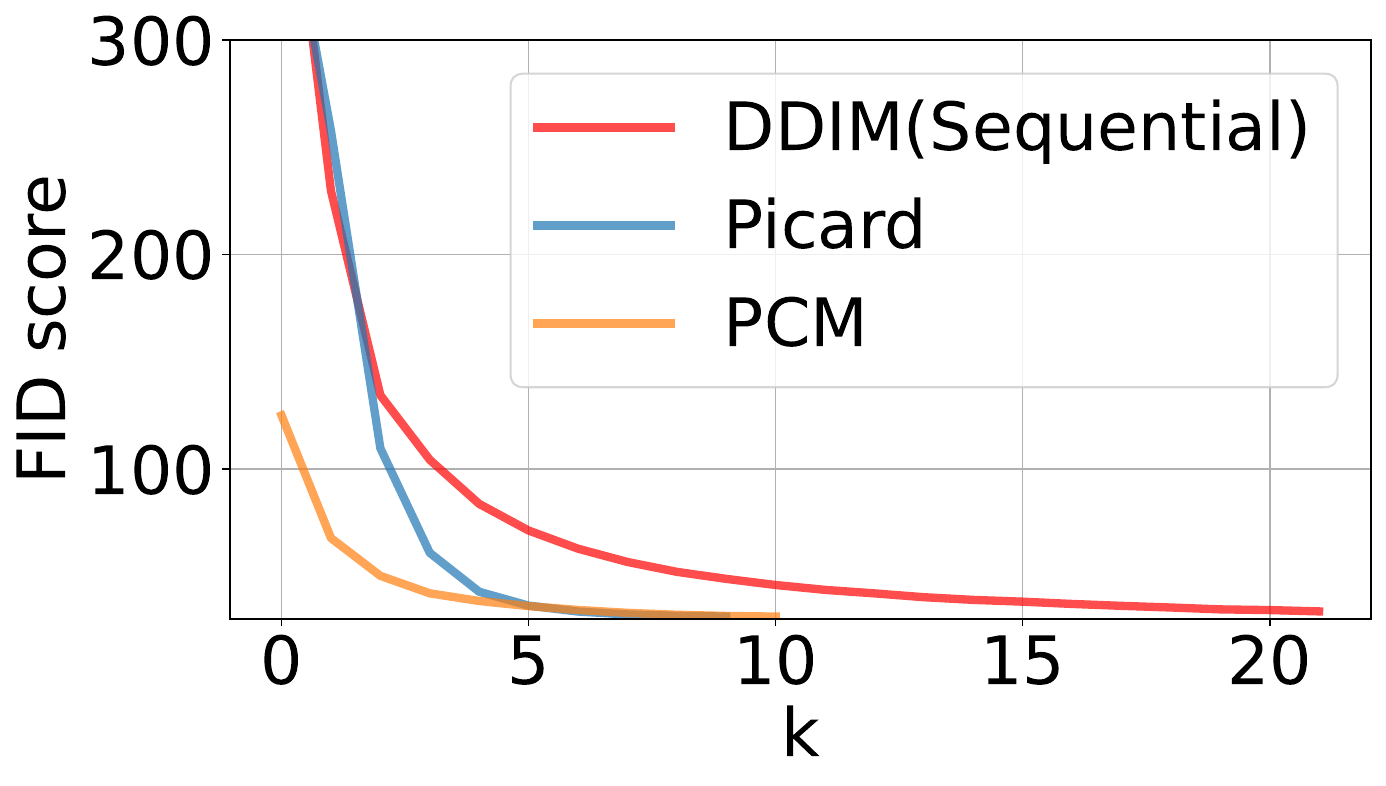}
                \caption{CelebA}
    \end{subfigure}
        \begin{subfigure}[b]{0.45\columnwidth}
                \includegraphics[width=\textwidth]{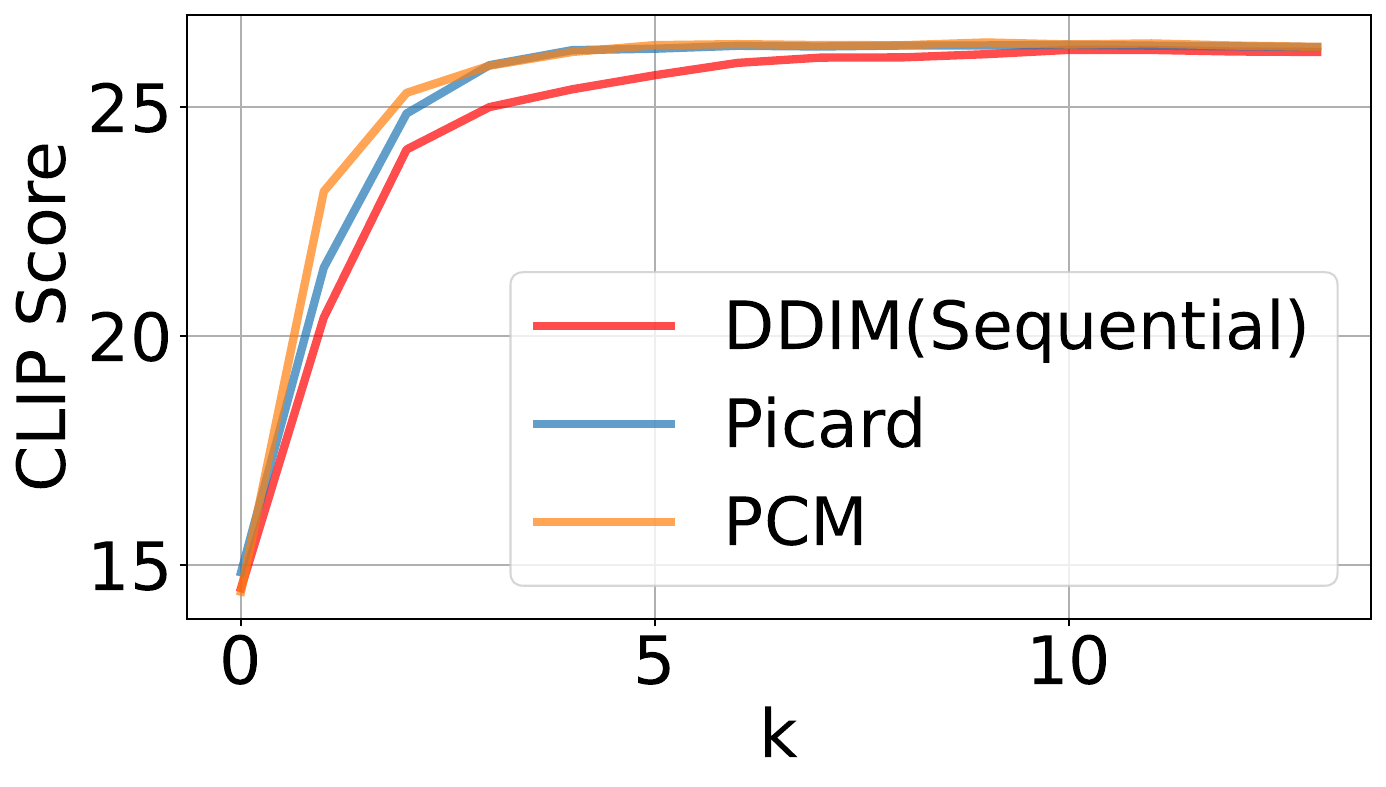}
                \caption{Stable Diffusion}
    \end{subfigure}
    \caption{Pareto front comparison of diffrent sampling methods, Sequeuntial, Picard and our PCM. Our PCM show consistent improvement in every iteration compared to oteher methods.}
    \label{fig:pareto_comp}
\end{figure}

In Fig. \ref{fig:pareto_comp},  we present a Pareto front comparison of Sequential(DDIM), Picard, and our proposed PCM, evaluated based on their difference from the ground truth (GT) over the same sequential iteration $k$. As illustrated in the figures, our PCM consistently outperforms all other methods given the same number of sequential iterations, achieving a complete Pareto front in both CelebA and Stable diffuison dataset.
  
\section{Limitation}
\label{sec:limitation}

In this study, we propose the Picard Consistency Model, which significantly accelerates the convergence speed of Picard Sampling. The primary limitation of our approach is that it requires training, which incurs a higher computational cost compared to existing training-free acceleration methods. However, our method does not require additional datasets, completes training within relatively few epochs. Also, since training is a one-time cost, PCM introduces no overhead during the inference. 
The second limitation of our approach is a challenge shared by all parallel inference methods: reducing latency necessitates increased computational resources and energy. However, considering the current trend in computing, which is moving toward maximizing parallelism, we believe that parallel computation costs will become less of a concern. Instead, reducing latency is likely to deliver greater value in practice.
\section{Additional Visualization Examples}
\label{sec:visual}

\begin{figure*}[t]
    \centering
        \begin{subfigure}[b]{\linewidth}
                \includegraphics[width=\textwidth]{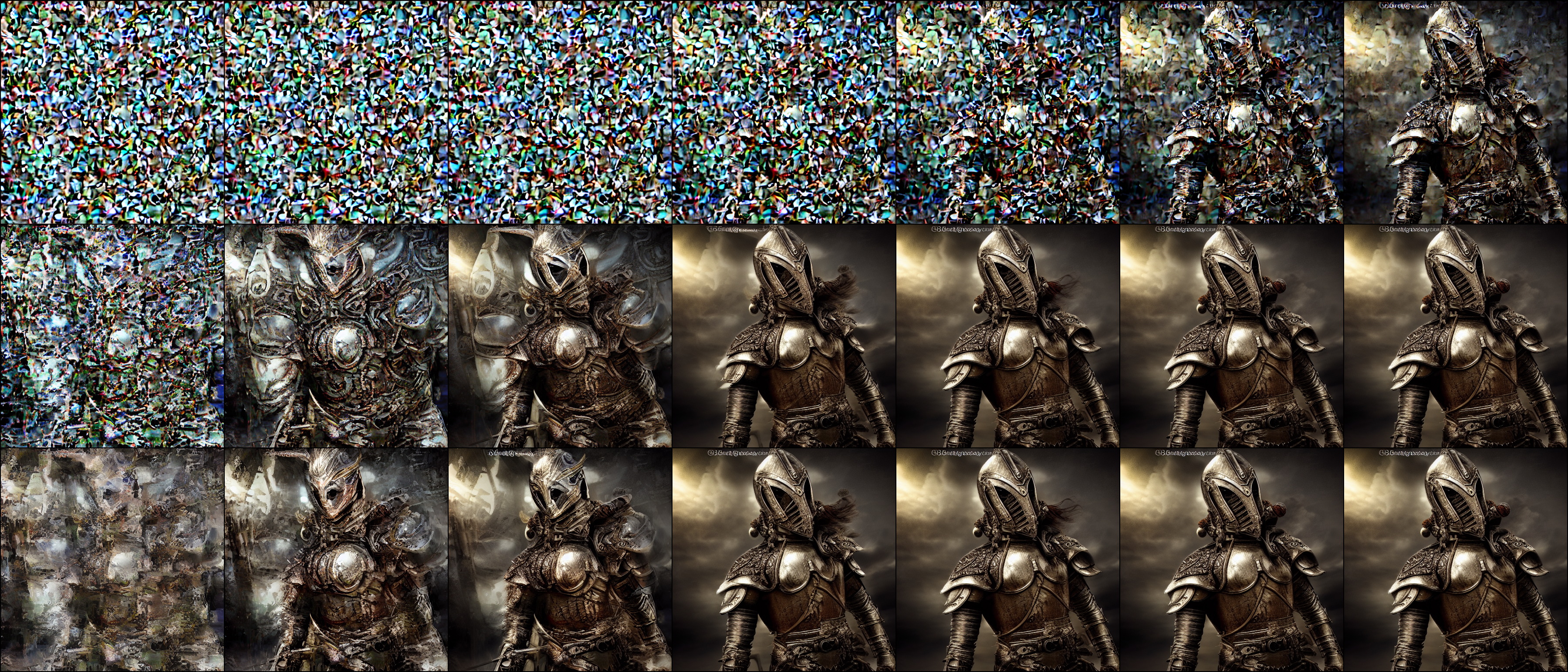}
    \end{subfigure}
    \caption{Stable Diffusion v1.4 Qualtitive comparison of DDIM(up-row), Picard (mid-row) and PCM(down-row) in same iteration $k$. The prompt is "a warrior in gleaming armor, standing on a battlefield, dramatic lighting, ultra realistic, intricate details, vivid, hdr, cinematic".
    }
    \label{fig:compare_sd}
\end{figure*}

\begin{figure*}[t]
    \centering
        \begin{subfigure}[b]{\linewidth}
                \includegraphics[width=\textwidth]{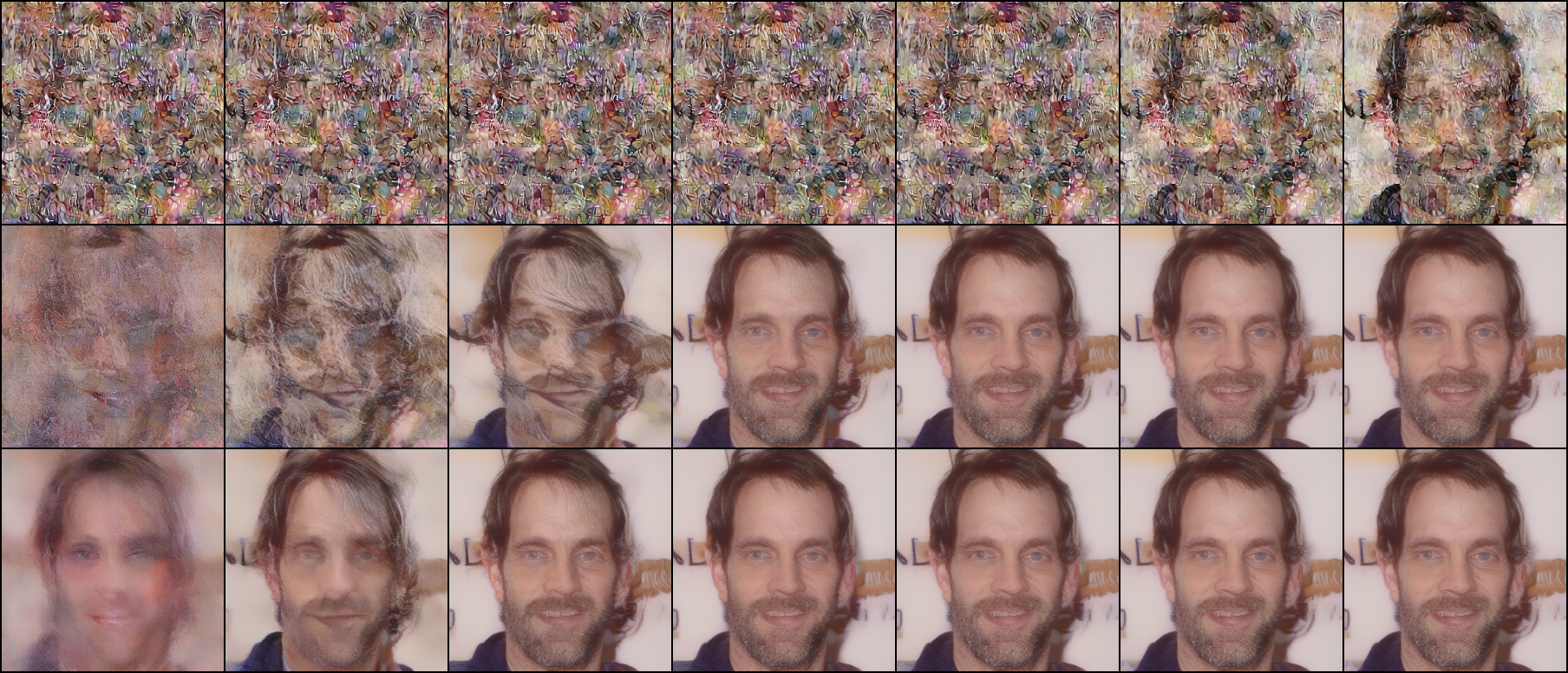}
    \end{subfigure}
    \caption{CelebA Qualtitive comparison of DDIM(up-row), Picard (mid-row) and PCM(down-row) in same iteration $k$.
    }
    \label{fig:compare_celeb}
\end{figure*}

In Fig. \ref{fig:compare_sd},\ref{fig:compare_celeb}, we provide additional visual examples for CelebA and Stable Diffusion, comparing the quality of image generation using DDIM (Sequential), Picard, and our proposed PCM. As shown in the figures, while DDIM and Picard converge slowly during the initial iterations and produce inaccurate images, our PCM generates plausible images earlier and converges more quickly.

\begin{figure*}[t]
    \centering
        \begin{subfigure}[b]{\linewidth}
                \includegraphics[width=\textwidth]{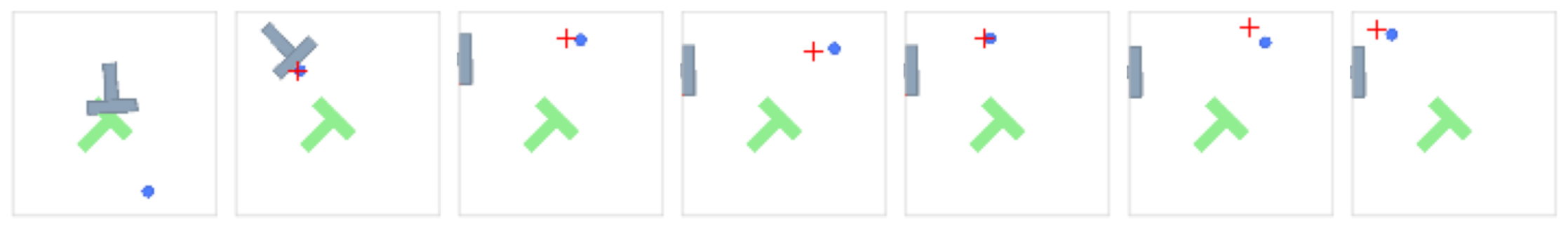}
    \end{subfigure}
     \begin{subfigure}[b]{\linewidth}
                \includegraphics[width=\textwidth]{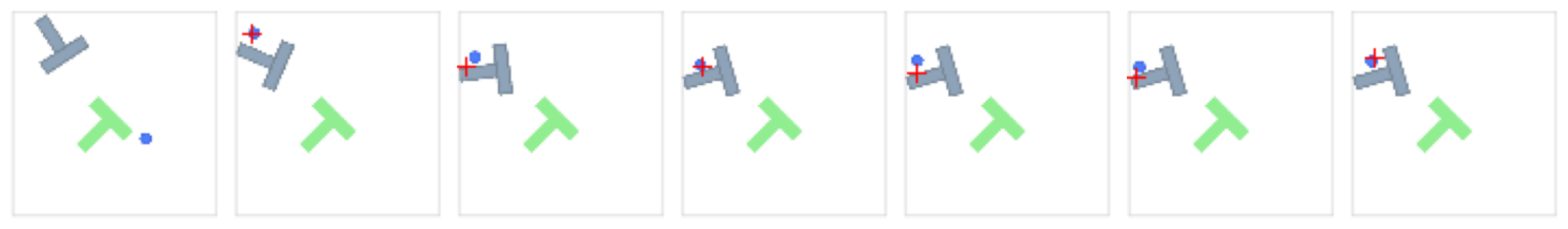}
    \end{subfigure}
     \begin{subfigure}[b]{\linewidth}
                \includegraphics[width=\textwidth]{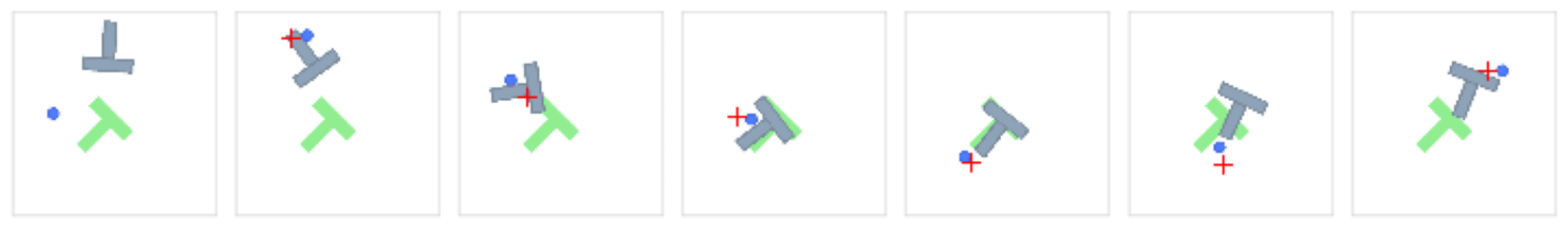}
    \end{subfigure}
    \caption{Generated action episode using Picard from $k=4$, we randomly sample 3 epsiodes using different random seeds.
    }
    \label{fig:pusht_picard}
\end{figure*}

\begin{figure*}[t]
    \centering
        \begin{subfigure}[b]{\linewidth}
                \includegraphics[width=\textwidth]{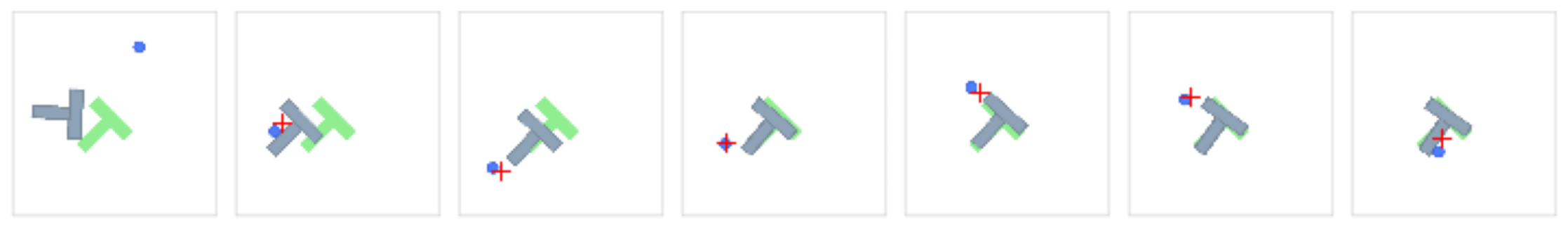}
    \end{subfigure}
     \begin{subfigure}[b]{\linewidth}
                \includegraphics[width=\textwidth]{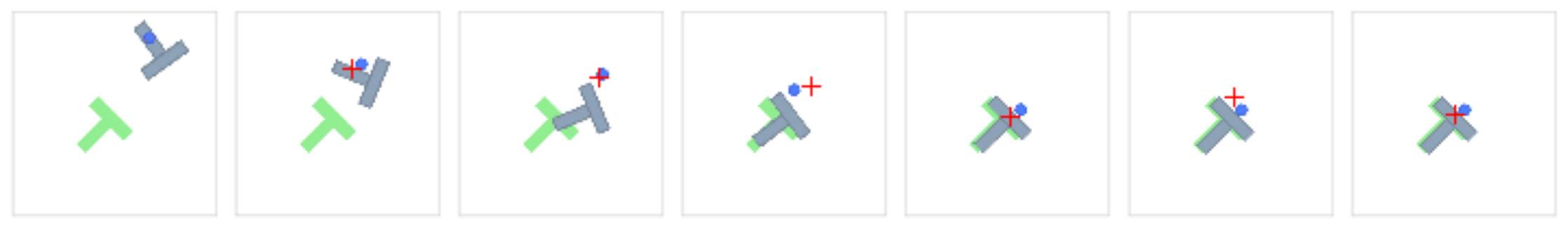}
    \end{subfigure}
     \begin{subfigure}[b]{\linewidth}
                \includegraphics[width=\textwidth]{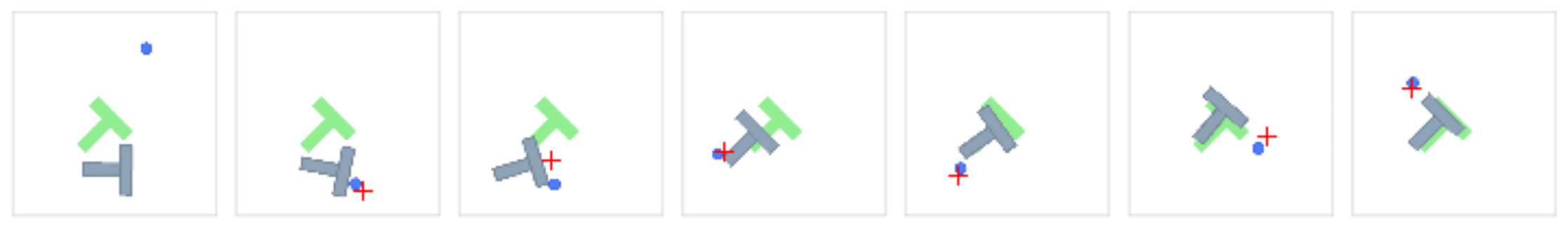}
    \end{subfigure}
    \caption{Generated action episode using PCM from $k=4$, we randomly sample 3 epsiodes using different random seeds.
    }
    \label{fig:pusht_pcm}
\end{figure*}

In Figures \ref{fig:pusht_picard} and \ref{fig:pusht_pcm}, we illustrate the action trajectories of Picard and our PCM using three different random seeds starting from $k=4$. As shown in the figures, while Picard fails to generate accurate motions in most cases, our PCM successfully produces correct motions in nearly all scenarios.

\end{document}